\documentclass[journal]{IEEEtran}
\usepackage{amsmath,amsfonts}
\usepackage{algpseudocode}
\usepackage{listings}
\usepackage{array}
\usepackage[caption=false,font=normalsize,labelfont=sf,textfont=sf]{subfig}
\usepackage{textcomp}
\usepackage{stfloats}
\usepackage{url}
\usepackage{verbatim}
\usepackage{graphicx}
\usepackage{cite}
\usepackage{hyperref}
\usepackage{xcolor}
\usepackage{listings}

\definecolor{codegreen}{rgb}{0,0.6,0}
\definecolor{codepurple}{rgb}{0.58,0,0.82}
\definecolor{backcolour}{rgb}{0.95,0.95,0.92}

\lstdefinestyle{mystyle}{
	backgroundcolor=\color{backcolour},   
	commentstyle=\color{codegreen},
	keywordstyle=\color{magenta},
	numberstyle=\footnotesize\color{gray},
	stringstyle=\color{codepurple},
	basicstyle=\ttfamily\small,
	breakatwhitespace=false,         
	breaklines=true,                 
	captionpos=b,                    
	keepspaces=true,                 
	numbers=left,                    
	numbersep=3pt,                  
	showspaces=false,                
	showstringspaces=false,
	showtabs=false,                  
	tabsize=2
}

\newcommand{\eg}{{\it e.g.}}
\newcommand{\ie}{{\it i.e.}}

\newcommand{\BEQ}{\begin{equation}}
\newcommand{\EEQ}{\end{equation}}
\newcommand{\BEAS}{\begin{eqnarray*}}
\newcommand{\EEAS}{\end{eqnarray*}}
\newcommand{\reals}{{\mbox{\bf R}}}

\newcommand{\pinit}{p^\mathrm{init}}
\newcommand{\pterm}{p^\mathrm{term}}

\newcommand{\conv}{\mathop{\bf conv}}

\usepackage{relsize} % Relsize package needed

\begin{document}

\title{Fast Path Planning\\ Through Large Collections of Safe Boxes}

\author{Tobia~Marcucci, Parth~Nobel, Russ~Tedrake, and Stephen~Boyd% <-this % stops a space
\thanks{T.~Marcucci  and R.~Tedrake are with the Department of Electrical Engineering and Computer Science, Massachusetts Institute of Technology,
Cambridge, MA 02139, USA.
P.~Nobel and S.~Boyd are with the Department of Electrical Engineering, Stanford University,
Stanford, CA 94305, USA.
Corresponding author is T. Marcucci: \texttt{tobiam@mit.edu}.
}% <-this % stops a space
%\thanks{Manuscript received MONTH DAY, YEAR; revised MONTH DAY, YEAR.}
}

\maketitle

\begin{abstract}
We present a fast algorithm for the design of smooth paths
(or trajectories) that are constrained to lie in a collection of axis-aligned boxes.
We consider the case where the number of these safe boxes is large, 
and basic preprocessing of them (such as finding their
intersections)
can be done offline.
At runtime we quickly generate a smooth path between given
initial and terminal positions.
Our algorithm designs trajectories that are guaranteed to be
safe at all times, and detects infeasibility whenever such a trajectory
does not exist.
Our algorithm is based on two subproblems that we can solve very efficiently:
finding a shortest path in a weighted graph, and solving
(multiple) convex optimal-control problems.
We demonstrate the proposed path planner on large-scale numerical examples,
and we provide an efficient open-source software 
implementation, \texttt{fastpathplanning}.
\end{abstract}


\section{Introduction}

Path planning is a problem at the core of almost any autonomous system.
Driverless cars, drones, autonomous aircraft, robot manipulators, and legged
robots are just a few examples of systems that rely on a path-planning algorithm
to navigate in their environment.
Path-planning problems can be challenging on many fronts.
The environment can be dynamic, \ie, change over time, or
uncertain because of noisy sensor
measurements~\cite{lavalle1997motion,fiorini1998motion,petti2005safe,du2011robot}.
Computation might be subject to strict real-time
requirements~\cite{frazzoli2002real,kuwata2009real,katrakazas2015real}.
Interactions between multiple robots without central
coordination can lead to game-theoretic
problems~\cite{sadigh2016planning,wang2017safety,liniger2019noncooperative,spica2020real}.
In this paper we consider problems
where a single smooth path needs to be found through
an environment that is fully known and static, but 
potentially very large and complicated to navigate through.
For example, this is the case for a drone inspecting 
an industrial plant or a mobile robot transporting packages in 
a large warehouse.

\begin{figure}
\centering
\includegraphics[width=\columnwidth]{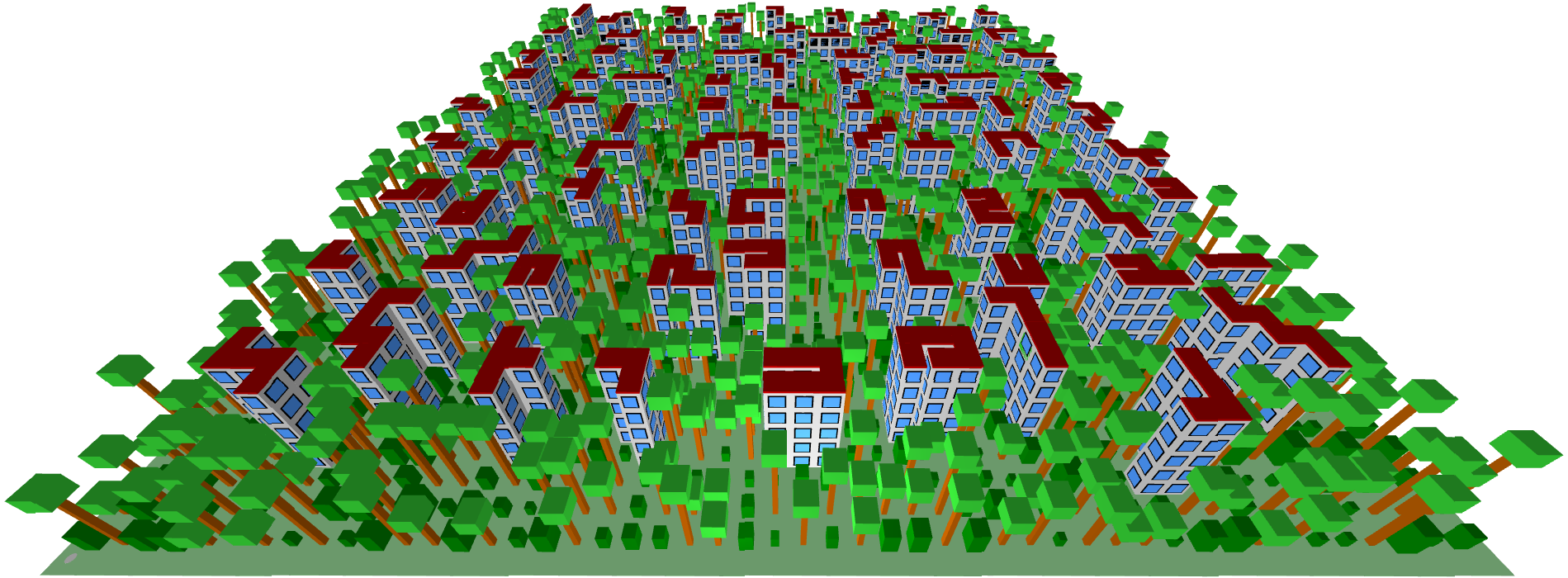} \\
\vspace{1mm}
\includegraphics[width=\columnwidth]{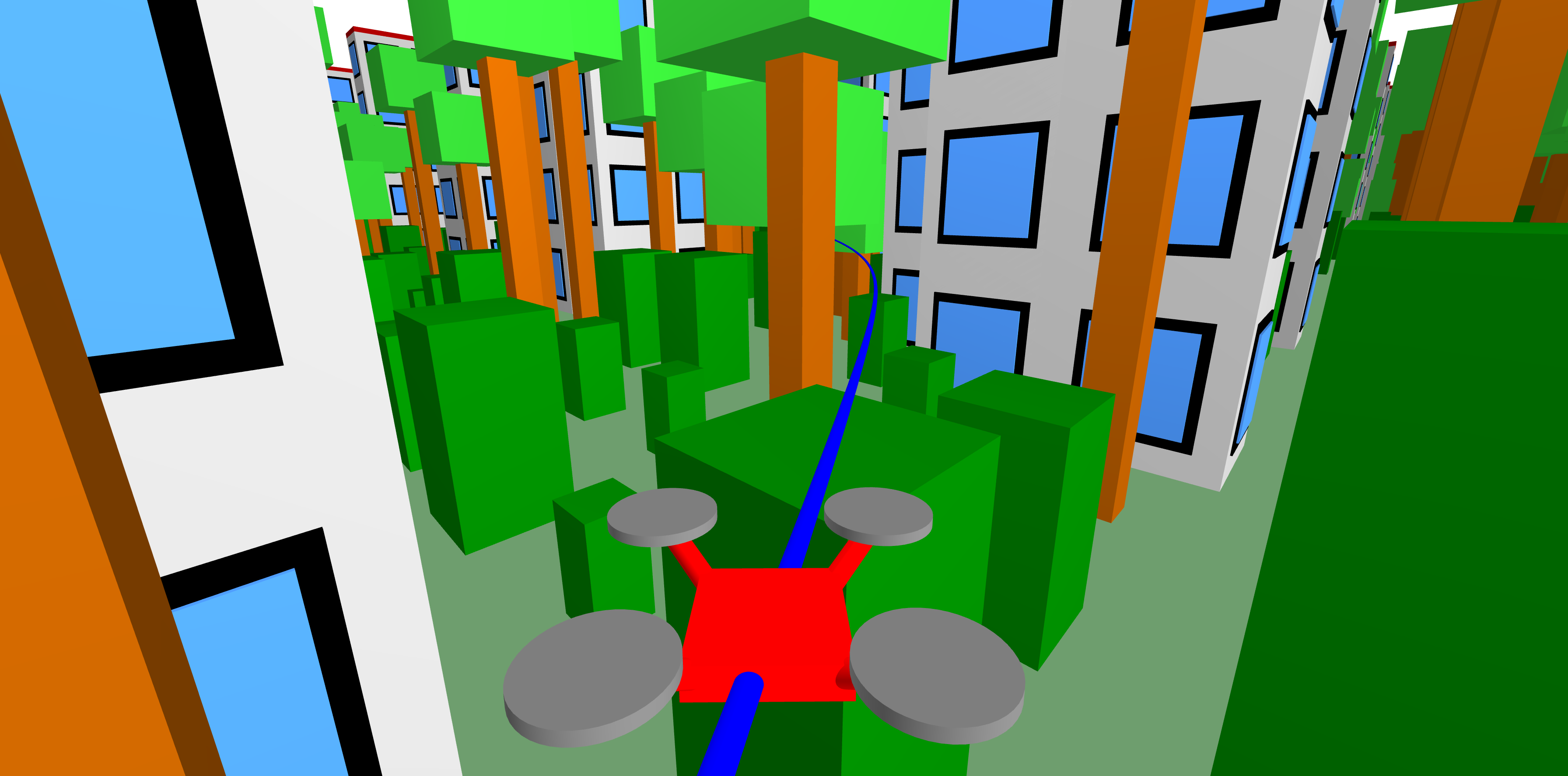}
\caption{
Path planning for a quadrotor flying through a simulated village.
\emph{Top.} The village, composed of buildings, trees, and bushes.
The free space is decomposed using more than ten thousand safe boxes.
\emph{Bottom.} A snapshot of the quadrotor flight.
The smooth path connects two opposite corners of the village and
is guaranteed to be collision free at all times.
The online planning time is only a few seconds.
}
\label{f-quad}
\end{figure}

Like previous methods~\cite{deits2015efficient,marcucci2023motion},
we assume that the environment is described as a
collection of safe sets, through which our system or
robot can move freely without colliding with obstacles.
Our problem is to find a smooth path
that is contained in the union of the safe sets, and
connects given initial and terminal points.
We consider the case where the safe sets are
axis-aligned boxes and large in number (thousands or tens of thousands).
Note that the decomposition of complex environments
into boxes can be approximate (conservative), and can
be computed using simple variations of existing
algorithms~\cite{lien2004approximate,ayanian10,ghosh2013fast,deits2015computing,werner2023approximating},
as well as methods tailored to kinematic
trees~\cite{amice2022finding,verghese2022configuration,dai2023certified,petersen2023growing}.
Focusing on box-shaped safe sets allows us to substantially accelerate
multiple parts of our algorithm.

Our path-planning method is composed of an offline and an online part.
In the offline preprocessing, we construct a graph
that stores the intersections
of the safe boxes
and solve a convex program to label the edges of
this graph with approximate distances.
These computations are done only once,
since the environment is static, and they require from
a fraction of a second to a few tens of
seconds, depending on the problem size.
In the online part we first use the graph constructed offline
to design a polygonal
curve of short length that connects the given initial and terminal points.
Then we solve a sequence of convex optimal-control problems
to transform the polygonal curve into a smooth path
that minimizes a given objective function.
The online runtimes of our algorithm are dominated by these
control problems, which, however, are solvable in a
time that increases only linearly with the number of boxes traversed
by the path~\cite{wang2009fast}.
This results in online planning times on the scale of a
hundredth of a second for medium-size problems
and of a second for very large problems.
Consider that, for a problem like the quadrotor flight in Fig.~\ref{f-quad},
existing techniques take a few seconds to find a path
through less than one hundred safe boxes~\cite{marcucci2023motion}.
Within the same time, our planner designs a path
through more than ten thousand boxes.

The proposed algorithm is \emph{complete}:
it always finds a smooth path connecting the initial and final positions
if such a path exists,
and it certifies infeasibility of the planning problem otherwise.
In addition, by using B\'ezier curves for the path parameterization,
our smooth trajectories are guaranteed to be safe at all times,
and not only at a finite number of sample points.
Our method is \emph{heuristic}:
although it designs paths that have typically low cost,
it is not guaranteed to solve the planning problem optimally,
or within a fixed percentage of the optimum.
The techniques of this paper are implemented in a companion
open-source package, \texttt{fastpathplanning}.

\subsection{Related work}

A wide variety of path-planning algorithms have been developed over the last
fifty years. An excellent overview of the techniques available in the
literature can be found in~\cite[Part~2]{lavalle2006planning}.
Here we review the methods that are most closely related to ours.

The closest approach to the one presented here
is GCS (graphs of convex sets) from~\cite{marcucci2023motion}. 
Similarly to our method, GCS designs smooth paths
through collections of safe sets that are preprocessed to form a graph.
Leveraging the optimization framework from~\cite{marcucci2021shortest},
it formulates a tight convex relaxation of the planning problem and it recovers
a collision-free trajectory using a cheap rounding strategy.
Thanks to this workflow, GCS also provides tight optimality bounds for
the trajectories it designs.
On the other hand, by trying to solve the planning problem through
a single convex program, GCS has a few limitations.
First, at present,
GCS does not efficiently handle costs or constraints
on the path acceleration and higher derivatives,
which are a central component of the problems analyzed in this paper.
Secondly, GCS does not scale to the very large numbers of safe sets
considered here.
The proposed algorithm is different in spirit from GCS:
it leverages fast graph search to heuristically solve
the discrete part of the planning problem and,
only at a later stage, it uses convex optimization to shape
the continuous path. This division sacrifices the optimality guarantees
but retains the algorithm completeness, and it allows us
to quickly find paths of low cost for planning problems of very large scale.

A natural approach for designing smooth paths that avoid
obstacles optimally is mixed-integer programming.
Earlier mixed-integer formulations dealt with
polyhedral obstacles, and used a binary variable for each facet of
each obstacle to enforce the constraint that a trajectory point is not in
collision~\cite{schouwenaars2001mixed,richards2002coordination,mellinger2012mixed}.
Conversely, the formulation from~\cite{deits2015efficient} leverages the
algorithm from~\cite{deits2015computing} to cover
(all or part of) the collision-free space with convex regions,
and ensures safety by forcing each trajectory segment
to lie entirely in at least one convex region.
This makes the mixed-integer program more efficient, since each
trajectory segment requires only one binary variable per safe set.
The path planning problem considered in~\cite{deits2015efficient}
is essentially the same as ours, but our algorithm
can solve dramatically larger problems in a fraction of the time
(see the comparison in \S\ref{s-mip}).

Two popular approaches for collision-free path planning are
local nonconvex optimization~\cite{augugliaro2012generation,schulman2014motion,liu2014solving,majumdar2017funnel,bonalli2019gusto,zhang2020optimization}
and sampling-based
algorithms~\cite{kavraki1996probabilistic,lavalle1998rapidly,karaman2011sampling}.
The former methods can handle
kinematic and dynamic constraints, but
suffer from local minima and can often fail
in finding a feasible trajectory if the environment has many obstacles.
Although multiple approaches have been proposed to mitigate this
issue~\cite{ratliff2009chomp,kalakrishnan2011stomp,hauser2016learning},
sampling-based algorithms are typically more reliable when the
environment is complex (in fact, they are \emph{probabilistically complete}).
However, sampling-based methods can struggle in high dimensions
and are less suitable for the design of smooth paths.
Similar to~\cite{marcucci2023motion}, the approach we
propose here can be thought of as a generalization of
sampling-based algorithms, where collision-free samples are
substituted with collision-free sets.
Instead of sampling the environment densely,
we fill it with large safe boxes.
This reduces the combinatorial complexity to the minimum and
allows us to plan through the open space using
efficient convex optimization~\cite{boyd2004convex}.

Decompositions of the environment into safe sets
(or cells) are also common in feedback motion planning.
There, a feedback plan is constructed by composing a
navigation function with a piecewise-smooth vector field:
the former decides the discrete transitions between the cells and
the latter causes all states in a cell to flow into the next
cell~\cite[\S8.4]{lavalle2006planning}.
In a similar fashion, the method in~\cite{belta2005discrete}
leverages discrete abstractions~\cite{alur2000discrete}
to generate provably correct control
policies for planar robots moving in polygonal environments.
In robust motion planning, funnels~\cite{tedrake2010lqr,majumdar2017funnel},
tubes~\cite{singh2017robust},
barrier functions~\cite{ames2016control,wang2017safety},
and positively invariant sets~\cite{weiss2014spacecraft,weiss2015safe,berntorp2019positive,danielson2020robust}
are frequently used to abstract away the continuous dynamics
and reduce the planning problem to a discrete search.
While similar in spirit to our algorithm, the methods presented in
those papers consider problems of different nature from our.
We do not aim to synthesize a feedback policy,
nor do we reason about dynamics and disturbances explicitly.
Our goal is to design safe smooth paths of low cost, and the challenge in
our problem is the environment complexity
(\ie, the number of safe boxes).

Lastly, in this paper we use B\'ezier curves to parameterize smooth paths.
These curves enjoy several properties that make them
particularly well suited for convex optimization,
and have been widely used in path planning and optimal control over the last fifteen 
years~\cite{choi2008path,flores2008real,lau2009kinodynamic,elbanhawi2015continuous,park2019fast,koolen2020balance,tordesillas2021faster,csomay2022multi,arslan2022adaptive,marcucci2023motion}.

\subsection{Outline}

This paper is organized as follows.
In \S\ref{s-pp} we state the path-planning problem and 
give a high-level overview of our algorithm.
The algorithm can be broken down into three parts,
one offline and two online.
The offline preprocessing, which does not use either the endpoints
of the path or the specific objective function, is 
described in \S\ref{s-preprocessing}.
The first online phase, illustrated in \S\ref{s-polygonal},
finds a polygonal curve of short length that is contained in the safe boxes
and connects the given path endpoints.
The second online phase, described in \S\ref{s-smooth},
transforms the polygonal curve into a safe smooth path of low cost.
In \S\ref{s-analysis} we summarize the main properties
of our path planner.
In \S\ref{s-experiments} we evaluate the performance
of our algorithm through multiple numerical experiments.
In conclusion, in \S\ref{s-extensions}, we describe
some extensions of our method to more general planning problems.

\section{Path Planning} \label{s-pp}

In this section we state the path-planning problem
and we describe at a high-level the components of our algorithm.

\subsection{Problem statement}

We consider the design of a smooth path in $\reals^d$ from a given initial point 
$\pinit\in \reals^d$ to a given terminal point 
$\pterm\in \reals^d$. 
We represent the path as the function $p:[0,T]\to \reals^d$, where $T$ is the time
taken to traverse the path.
In addition to the initial and terminal point constraints,
\[
p(0)= \pinit, \qquad
p(T)= \pterm,
\]
we require that the path stay in a given set
$\mathcal S \subset \reals^d$ of safe points:
\[
p(t) \in \mathcal S, \quad t\in[0,T].
\]
We assume that the safe set $\mathcal S$ is a union of $K$ axis-aligned boxes,
\[
\mathcal S = \bigcup_{k=1}^K \mathcal B_k,
\]
with 
\[
\mathcal B_k = \{x \in \reals^d \mid l_k \leq x \leq u_k\}, \quad k=1, \ldots, K.
\]
Here the inequalities are elementwise, and the box bounds satisfy
$l_k \leq u_k$ for $k=1,\ldots, K$.

We considers paths with $D$ continuous derivatives, and we take
our objective to be a weighted sum of the squared $L_2$ norm
of these derivatives,
\BEQ
\label{e-obj}
J = \sum_{i=1}^D \alpha_i
\int_0^T \| p^{(i)} (t) \|_2^2 \; dt,
\EEQ
where $p^{(i)}$ denotes the $i$th derivative of $p$, and $\alpha_i$
are nonnegative weights.

The path-planning problem is
\BEQ \label{e-ppp}
\begin{array}{ll}
\mbox{minimize} & J \\
\mbox{subject to}
& p(0) = \pinit, \quad p(T) = \pterm, \\
& p(t) \in \mathcal S, \quad  t \in [0, T].
\end{array}
\EEQ
The optimization variable is the path $p$.
The problem data are the objective weights $\alpha_i$,
the final time $T$, the initial and terminal points 
$\pinit$ and $\pterm$, and
the safe set $\mathcal S$ (specified by the box bounds
$l_k$ and $u_k$).
This statement includes only the essential
components of a path-planning problem.
For example, here we specify the initial and terminal positions,
but do not constrain the initial and terminal derivatives.
In \S\ref{s-extensions} we will discuss multiple of these
simple extensions, and highlight the necessary modifications to our method.

The path-planning problem~\eqref{e-ppp} is infinite dimensional, but we will 
restrict candidate paths to piecewise B\'ezier curves, which are 
parameterized by a finite set of control points.

\subsection{Safety map}

Problem~\eqref{e-ppp} has convex quadratic objective,
two linear equality constraints, and the safety constraint, which,
in general, is not convex.
The safety constraint is an infinite collection of
disjunctive constraints,
that force the point $p(t)$, for each $t \in [0,T]$, to lie in at least
one of the boxes $\mathcal B_k$.
Ensuring safety of a path $p$ is then equivalent to
finding a function $s: [0,T] \to \{1, \ldots, K\}$ such that
\[
p(t) \in \mathcal B_{s(t)}, \quad t\in [0,T].
\]
The value $s(t)\in \{1, \ldots, K\}$ represents the choice of 
a safe box for the path at time $t$,
and the overall function $s$ can be
thought of as a \emph{safety map} for our path.

Our safety maps will have a finite number of transitions, \ie, 
will be of the form
\BEQ \label{e-safety-map}
s(t) = \begin{cases}
	s_1 & t \in [t_0, t_1] \\
	s_2 & t \in (t_1, t_2] \\
	\vdots \\
	s_N & t \in (t_{N-1}, t_N],
\end{cases}
\EEQ
where $0 = t_0< t_1 < \cdots < t_N = T$.
We will refer to $s_1, \ldots, s_N$ as the \emph{box sequence}
of the safety map $s$, and to
$T_1 = t_1 - t_0,\ldots, T_N = t_N - t_{N-1}$
as the \emph{traversal times}.

In terms of the safety map, the path-planning problem is
\BEQ \label{e-ppp2}
\begin{array}{ll}
\mbox{minimize} & J \\
\mbox{subject to}
& p(0) = \pinit, \quad p(T) = \pterm, \\
& l_{s(t)} \leq p(t) \leq u_{s(t)},
\quad t\in [0,T],
\end{array}
\EEQ
where the variables are the path $p$ and the safety map $s$.
We observe that if the box sequence $s_1, \ldots, s_N$ is fixed,
problem~\eqref{e-ppp2} reduces to a nonconvex but continuous
optimal-control problem, with the path $p$ and the traversal times
$T_1, \ldots, T_N$ as decision variables.
If we also fix the traversal times, then the safety map is
entirely specified, and problem~\eqref{e-ppp2} becomes a convex optimal-control problem
with quadratic objective and linear constraints.

\subsection{Feasibility}
\label{s-feas}

We will say that a safety map is \emph{feasible} if it satisfies 
\BEQ
\label{e-feasible-safety-map}
\begin{array}{l}
\pinit \in \mathcal B_{s_1}, \quad \pterm \in \mathcal B_{s_N}, \\
\mathcal B_{s_j} \cap \mathcal B_{s_{j+1}} \neq \emptyset, \quad
j = 1, \ldots, N-1.
\end{array}
\EEQ
The first condition says that the first and last boxes in
the box sequence cover the initial and terminal points, respectively.
The second condition requires that every two consecutive boxes intersect;
thus the box sequence can be traversed by a continuous path.

Importantly, the path-planning problem~\eqref{e-ppp2} is feasible if and only if a
feasible safety map exists.
To see this, note that if a path $p$ and a safety map $s$ are
feasible for~\eqref{e-ppp2}, then the safety map must satisfy both
conditions in~\eqref{e-feasible-safety-map}.
(In particular, the second condition
follows from the continuity of $p$, which ensures that
$p(t_j) \in \mathcal B_{s_j} \cap \mathcal B_{s_{j+1}}$
for all $j=1, \ldots, N-1$.)
For the other direction, suppose a safety map is feasible,
and let $p_j \in \mathcal B_{s_j} \cap \mathcal B_{s_{j+1}}$
for $j=1, \ldots, N-1$.
Then the polygonal curve with nodes
$\pinit = p_0$, $p_1, \ldots, p_{N-1}, p_N=\pterm$ is entirely contained in the
safe set $\mathcal S$. 
Through the following steps,
we construct a path $p$ that has $D$ continuous derivatives,
and moves along the polygonal curve (and so is safe).
We select any times $0=t_0 < t_1 < \cdots < t_N=T$.
We choose any smooth time
parameterization of the polygonal curve that satisfies
the interpolation conditions $p(t_j)=p_j$ for $j=0, \ldots, N$,
as well as the derivative constraints
$p^{(i)}(t_j)=0$ for $i=1, \ldots, D$ and $j=1,\ldots, N-1$.
While the polygonal curve has kinks, the path $p$ is 
differentiable $D$ times since it comes to a full stop at each kink.
By pairing this path with the feasible safety map, we have a feasible
solution of problem~\eqref{e-ppp2}.

\subsection{Method outline}
\label{s-general-method}

We give here a high-level description of the
three phases in our path-planning algorithm, 
with the details illustrated in future sections.

\paragraph*{Offline preprocessing}

The offline preprocessing uses only the safe set  $\mathcal S$, \ie, the
safe boxes $\mathcal B_1, \ldots, \mathcal B_K$.
In this phase we construct a \emph{line graph} $G$
whose vertices correspond to points in the 
intersection of two boxes, and whose edges connect pairs of points
that lie in the same box.
When considered as a subset of $\reals^d$,
this graph lies entirely in the safe set,
and it can be used to quickly design safe polygonal curves that
connect given initial and terminal points.
The points associated with the vertices are called
\emph{representative points}, and are optimized to minimize
the total Euclidean length of the edges in the line graph.
This serves as a heuristic to reduce the length of the polygonal curves.

\medskip
\begin{algorithmic}[1]
\Procedure {Offline Preprocessing}{}
\State compute intersections of safe boxes $\mathcal B_1, \ldots, \mathcal B_K$
\State construct line graph $G$
\State optimize representative points
\EndProcedure
\end{algorithmic}

\paragraph*{Polygonal phase}
Here we find a polygonal curve $\mathcal C$ that connects
$\pinit$ to $\pterm$,
is entirely contained in the safe set $\mathcal S$, and has small length.
The curve is initialized by solving a shortest-path problem in
the line graph constructed offline.
Then it is shortened through an iterative process, where we
alternate between minimizing the curve length for a fixed
box sequence,
and updating the box sequence for a fixed
polygonal curve.

\medskip
\begin{algorithmic}[1]
\Procedure {Polygonal Phase}{}
\State connect $\pinit$ to $\pterm$ with safe polygonal curve $\mathcal C$ 
\While {not converged}
\State fix box sequence $s_1, \ldots, s_N$ and shorten curve
\State fix curve and improve box sequence
\EndWhile
\EndProcedure
\end{algorithmic}

\paragraph*{Smooth phase}

In this phase we freeze the box sequence $s_1, \ldots, s_N$
identified in the polygonal phase,
and traversed by the curve $\mathcal C$.
As observed above, this reduces
problem~\eqref{e-ppp2} to a continuous
but nonconvex optimal-control problem.
To solve this control problem, we first use a simple
heuristic to estimate
initial traversal times $T_1, \ldots, T_N$.
Then we alternate between
two convex optimal-control problems.
In the first, we fix the traversal times (thus we specify the whole
safety map $s$) and optimize the shape of the path.
In the second, we attempt to
improve the traversal times by solving
a local convex approximation of the
nonconvex optimal-control problem.

\medskip
\begin{algorithmic}[1]
\Procedure {Smooth Phase}{}
\State fix box sequence $s_1, \ldots, s_N$
\State estimate traversal times $T_1, \ldots, T_N$
\While {not converged}
\State fix traversal times and optimize path $p$
\State
attempt improvement of traversal times
\EndWhile
\EndProcedure
\end{algorithmic}

\section{Offline Preprocessing}
\label{s-preprocessing}

In this section we describe the offline part of our algorithm.
The steps below are also illustrated through a simple example at
the end of the section.

\subsection{Line graph}

We start by computing the line graph
associated with the safe boxes.
The vertices of this graph are pairs of safe boxes that intersect,
and the edges connect pairs of intersections that share a box.
Formally, the line graph is an undirected graph
$G=(\mathcal V, \mathcal E)$ with vertices
\[
\mathcal V = \{ \{k, l\} \subseteq \{1, \ldots, K\} \mid
\mathcal B_k \cap \mathcal B_l \neq \emptyset, \ k \neq l\},
\]
and edges
\[
\mathcal E = \{\{v, w\} \subseteq \mathcal V \mid
v \cap w \neq \emptyset, \ v \neq w \}.
\]
The name line graph is motivated by the fact that
$G$ can be equivalently defined as the line graph
of the \emph{intersection graph} of 
our collection of boxes.

The line graph allows us to efficiently construct polygonal
curves that are guaranteed to be safe.
Consider a path in the line graph.
For each vertex in this path, choose a point in $\reals^d$
in the corresponding box intersection.
Then form the polygonal curve passing through these points.
Each line segment in this curve is associated with an edge in the line graph,
and therefore with a safe box.
By construction, this safe box contains the line segment entirely.
It follows that the whole polygonal curve is safe.

Since computing the intersection of two boxes
is a trivial operation, we can construct the line graph
$G$ very efficiently,
even when the number $K$ of boxes is very large.
Our implementation
is based on the technique from~\cite[\S2]{zomorodian2000fast}.

\subsection{Representative points}

\begin{figure*}
\centering
\includegraphics[width=.23\textwidth]{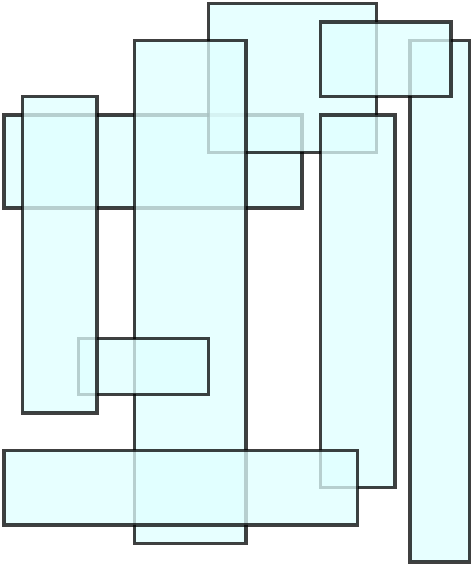} \quad
\includegraphics[width=.23\textwidth]{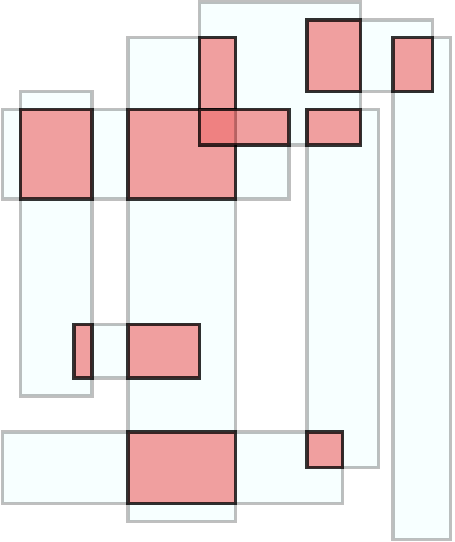} \quad
\includegraphics[width=.23\textwidth]{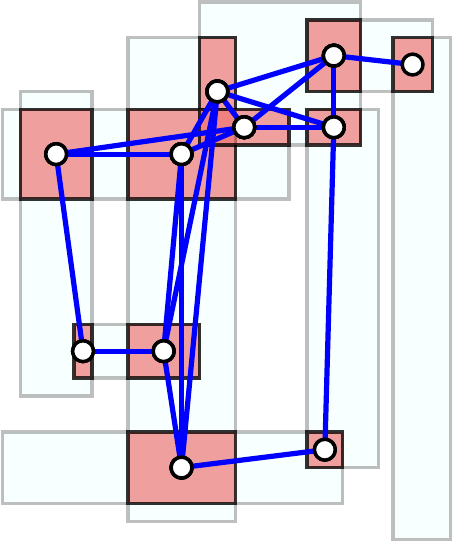} \quad
\includegraphics[width=.23\textwidth]{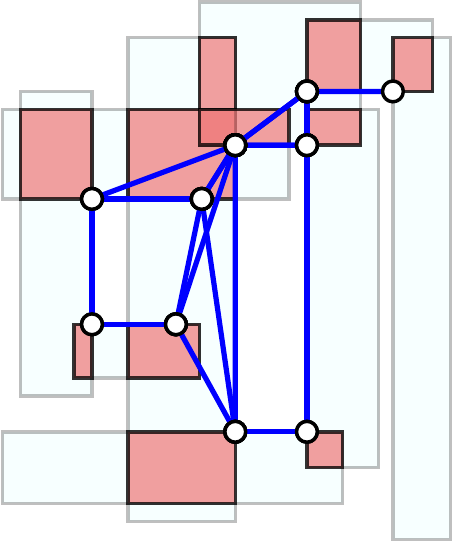}
\caption{
Offline preprocessing of the safe boxes.
\emph{Left.} Safe boxes. 
\emph{Center left.} Pairwise intersections of the safe boxes.
\emph{Center right.} Line graph, with vertices in the box
intersections and edges connecting intersections that
share a box.
\emph{Right.} Line graph with optimized representative points.}
\label{f-line-graph-and-pts}
\end{figure*}

Our next step is to choose a representative point for each 
vertex of the line graph, \ie, for each pair of intersecting boxes.
As a heuristic method to shorten
the polygonal curves constructed as described above,
we position these points close to their neighbors in the line graph.
More formally, denoting with $x_v \in \reals^d$ the representative point
of vertex $v \in \mathcal V$, we minimize the sum
of the Euclidean distances between all pairs of
representative points that are connected by an edge:
\BEQ\label{e-rep-pts-opt}
\begin{array}{ll}
\mbox{minimize} & \sum_{\{v, w\} \in \mathcal E} \|x_v-x_w\|_2 \\
\mbox{subject to} &
x_v \in \mathcal B_k \cap  \mathcal B_l, \quad v = \{k, l\} \in \mathcal V.
\end{array}
\EEQ
Here the variables are the representative points $x_v$, $v \in \mathcal V$.
Each of these points is constrained in the
corresponding box intersection, which is itself an axis-aligned box.
This is a convex optimization problem
that can be represented as a second-order
cone program (SOCP) and efficiently solved \cite[\S4.4.2]{boyd2004convex},~\cite{lobo1998applications}.

After optimizing the position of the representative points
$x_v$ as in~\eqref{e-rep-pts-opt}, each edge $\{v, w\}$
of the line graph is assigned the weight $\|x_v-x_w\|_2$.

\subsection{Example}
\label{s-example1}

We illustrate the offline preprocessing on a small problem
that will serve as a running example throughout the paper.
This problem has $K=9$ safe boxes in $d=2$ dimensions and is
depicted in Fig.~\ref{f-line-graph-and-pts}.
The left figure shows the safe boxes,
and the center left figure shows their intersections (with some overlapping 
when more than two boxes intersect).
These intersections correspond to the $|\mathcal V|=11$ vertices of the line graph.
In the center right figure, we show the $|\mathcal E|=20$ edges of the
line graph as line segments connecting the centers of the box intersections.
The right figure shows the optimized representative points, 
which minimize the total Euclidean distance over the edges of the 
line graph, \ie, a solution of~\eqref{e-rep-pts-opt}.
Note that some of the $20$ edges overlap in this figure.
Observe also that the line graph, considered as a subset of $\reals^2$, is entirely
contained in the safe set, since each edge is in at least one safe box.

\section{Polygonal Phase}
\label{s-polygonal}

We now describe the first online phase of our algorithm,
where we design a safe polygonal curve $\mathcal C$
of short length that connects $\pinit$ to $\pterm$.
An illustration of the steps below can be found at the end of the
section, where we continue our running example.

\subsection{Shortest-path problem}
\label{s-shortest-path}

We use the line graph $G$ to initialize the polygonal curve $\mathcal C$.
We augment the line graph with two new vertices with representative 
points $\pinit$ and $\pterm$.
An edge is added between $\pinit$ and all the intersections
of safe boxes that contain $\pinit$, \ie, all the vertices
$\{k,l\} \in \mathcal V$ such that $\pinit \in \mathcal B_k$ or
$\pinit \in \mathcal B_l$.
An analogous operation is done for $\pterm$.
As for the other edges in the line graph, these
new edges are assigned a weight equal
to the Euclidean distance between the representative points that
they connect.
We then find a shortest path from
the initial point to the terminal point, and recover
an initial polygonal curve $\mathcal C$
by connecting the representative points along this path.
As noted above, this curve is safe because each of its
segments is contained in a safe box.

This shortest-path step determines whether
or not our path-planning problem is
feasible.  If there is no path in the augmented line graph between
the vertices associated with $\pinit$ and 
$\pterm$, then the path-planning problem~\eqref{e-ppp2} is infeasible.
Conversely, if there is a path between these two vertices,
then the path-planning problem is feasible
since a feasible trajectory can be constructed as in \S\ref{s-feas}.

The problem of identifying all the safe boxes
that contain the initial and terminal points
is known as \emph{stabbing problem} and,
given the precomputations done to construct
the line graph, it takes negligible time~\cite{zomorodian2000fast}.
Using an optimized implementation of Dijkstra's algorithm (\eg, the one provided
by \texttt{scipy}~\cite{scipy2020}), finding a shortest path is 
also very fast.

\subsection{Shortening of the polygonal curve}
\label{s-shortening}

Thanks to the optimization of the representative points
in~\eqref{e-rep-pts-opt}, our initial polygonal curve $\mathcal C$
is typically short.
However, the online knowledge of the initial and terminal points,
which were unknown during the preprocessing, can allow us to
shorten this curve further.
This is done iteratively: we alternate between solving a
convex program that minimizes the curve length
for a fixed box sequence, and improving the box sequence
for a fixed polygonal curve. 

\paragraph*{Optimization of the polygonal curve}
Denote with $\mathcal C$ the curve at the current iteration (initialized
with the solution of the shortest-path problem).
Let $N$ be the number of segments in $\mathcal C$
and $y_0, \ldots, y_N \in \reals^d$ be the curve nodes,
with $y_0 = \pinit$ and $y_N = \pterm$.
For $j=1, \ldots, N$, let also $s_j$ be the index of
the safe box that covers
the line segment between $y_{j-1}$ and $y_j$.
We fix the box sequence $s_1, \ldots, s_N$ traversed by the current curve, 
and we optimize the position of the nodes $y_j$ so that the
length of the curve is minimized.
This leads to the problem
\BEQ \label{e-shortening}
\begin{array}{ll}
\mbox{minimize} & \sum_{j=1}^N \|y_j - y_{j-1}\|_2 \\
\mbox{subject to}
& y_0 = \pinit, \quad y_N = \pterm, \\
& y_j \in \mathcal B_{s_j} \cap \mathcal B_{s_{j+1}}, \quad j=1, \ldots, N - 1,
\end{array}
\EEQ
with variables $y_0, \ldots, y_N$.
This is a small SOCP with banded constraints that can be solved
very efficiently,
in time that is only linear in the number $N$ of segments~\cite{wang2009fast}.

The optimal nodes from~\eqref{e-shortening} define our
new curve $\mathcal C$.
We will assume that these nodes are distinct,
since if two nodes coincide we can always eliminate one.

\begin{figure}
\centering
\includegraphics[width=.75\columnwidth]{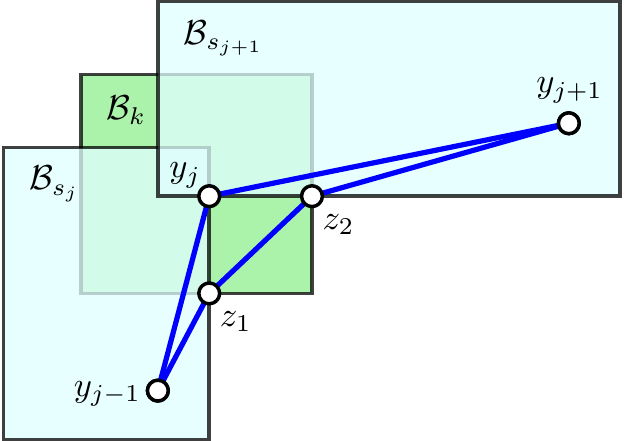}
\caption{Shortening of the polygonal curve $\mathcal C$ through
the insertion of a new box ($\mathcal B_k$ in green) in the current box sequence.}
\label{f-shortening}
\end{figure}

\paragraph*{Improvement of the box sequence}
After solving problem~\eqref{e-shortening}, the nodes
$y_0, \ldots, y_N$ minimize the curve length for the given box
sequence $s_1, \ldots, s_N$.
However, as Fig.~\ref{f-shortening} illustrates,
the insertion of a new box can potentially
give us room to further shorten our polygonal curve.
In our second step, we seek a new sequence of boxes that
contains the current curve and is guaranteed
to yield a length decrease.
Since our safe sets are boxes,
this step will be extremely quick.

For all $j=1,\ldots, N-1$, we solve a stabbing problem
to find all the boxes $\mathcal B_k$ that contain
the curve node $y_j$.
Then we consider inserting the index $k$ between $s_j$ and $s_{j+1}$
in our box sequence.
As shown in Fig.~\ref{f-shortening},
this insertion leads to a new instance of problem~\eqref{e-shortening}
where the variable $y_j$ is replaced by two variables:
\[
z_1 \in \mathcal B_{s_j} \cap \mathcal B_k, \qquad
z_2 \in \mathcal B_k \cap \mathcal B_{s_{j+1}}.
\]
Choosing $z_1 = z_2 = y_j$ gives us a feasible solution of this
new instance of~\eqref{e-shortening}, and does not change the
length of our curve $\mathcal C$.
Therefore the insertion of $\mathcal B_k$ leads to
a shorter curve if and only if this feasible solution is not optimal.

We fix $z_1 = z_2 = y_j$
and check if the optimality conditions of the
new instance of~\eqref{e-shortening} can be satisfied.
As explained in \S\ref{s-box-update},
this check reduces to finding a vector $\lambda \in \reals^d$ that
satisfies the following inequalities:
\BEQ \label{e-box-update}
\begin{array}{lll}
\|\lambda\|_2 \leq 1,
& L_1 (\lambda - \lambda_1) \geq 0,
& L_2 (\lambda - \lambda_2) \leq 0, \\
& U_1 (\lambda - \lambda_1) \leq 0,
& U_2 (\lambda - \lambda_2) \geq 0.
\end{array}
\EEQ
Here the vectors $\lambda_1, \; \lambda_2 \in \reals^d$ are fixed and given by
\[
\lambda_1 = \frac{y_j - y_{j-1}}{\| y_j - y_{j-1} \|_2}, \qquad
\lambda_2 = \frac{y_{j+1} - y_j}{\| y_{j+1} - y_j \|_2}.
\]
The matrices $L_1$ and $U_1$ select the indices of the inactive inequalities
in the box constraint $y_j \in \mathcal B_{s_j} \cap  \mathcal B_k$
($L_1$ for the lower bounds and $U_1$ for the upper bounds).
Similarly, $L_2$ and $U_2$ select the inactive inequalities
in $y_j \in \mathcal B_k \cap  \mathcal B_{s_{j+1}}$.

Checking if the inequalities in~\eqref{e-box-update} are
satisfiable is very quick.
In fact, since $L_1$, $L_2$, $U_1$, 
and $U_2$ are selection matrices, the corresponding inequalities
in~\eqref{e-box-update} simply impose
bounds on a subset of the entries of $\lambda$.
We express these bounds as $c_1 \leq \lambda \leq c_2$,
for two suitable vectors $c_1 \in (\reals \cup \{- \infty\})^d$ and
$c_2 \in (\reals \cup \{\infty\})^d$.
Then the vector $\lambda$ of minimum Euclidean norm that
lies within these bounds can be computed as
\BEQ \label{e-lambda}
\lambda^\star = \min \{c_2, \max \{c_1, 0\}\},
\EEQ
where the minimum and maximum are elementwise.
We conclude that $\lambda^\star$ has norm greater
than one if and only if the system of inequalities~\eqref{e-box-update}
has no solution, which is equivalent to
the insertion of the box $\mathcal B_k$
shortening our curve $\mathcal C$.

For each index $j=1,\ldots, N-1$ such that the norm of
$\lambda^\star$ is greater than one, 
we insert a new box in our sequence.
If multiple boxes satisfy this condition for the same
index $j$, we select one for which the norm of
$\lambda^\star$ is largest
(this heuristic is motivated in \S\ref{s-box-update}).
After updating the box sequence, we optimize the curve $\mathcal C$
by solving the new instance of problem~\eqref{e-shortening}.
This process is iterated until the condition above fails for every
curve node $j$ and every box $k$.

\subsection{Example}

\begin{figure*}
\centering
\includegraphics[width=.23\textwidth]{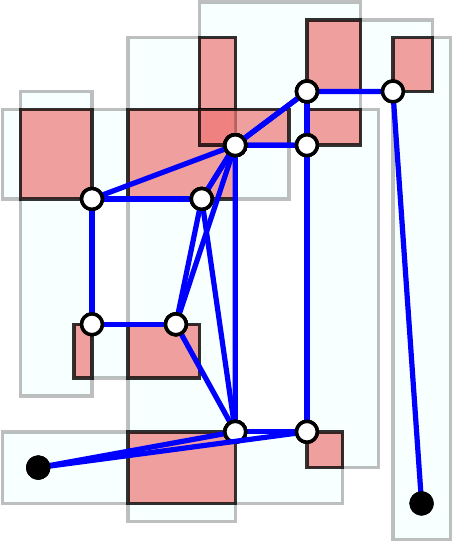} \quad
\includegraphics[width=.23\textwidth]{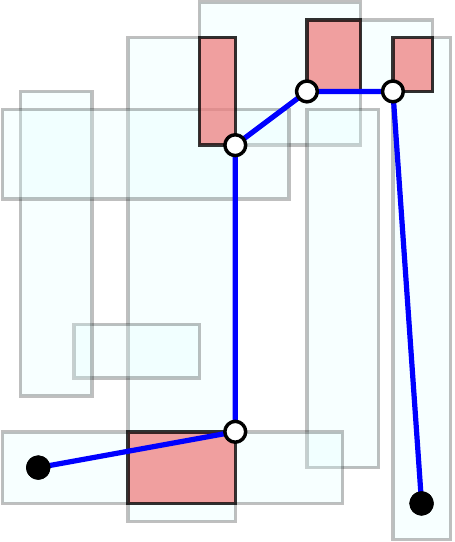} \quad
\includegraphics[width=.23\textwidth]{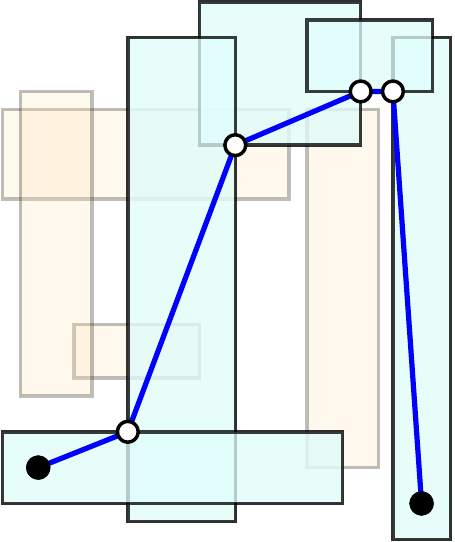} \quad
\includegraphics[width=.23\textwidth]{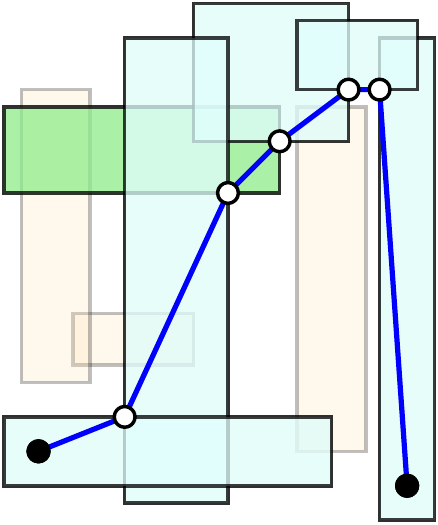}
\caption{
Polygonal phase of the algorithm.
\emph{Left.} Line graph augmented with $\pinit$ and $\pterm$, shown
as black disks.
\emph{Center left.} Shortest path from $\pinit$ to $\pterm$.
\emph{Center right.} The safe box sequence is fixed and the polygonal curve is 
shortened via convex optimization.
\emph{Right.} A new box (shown in green) is inserted in the sequence and the curve 
is shortened a second time.
Since no further shortening is possible, the polygonal phase converges in
one iteration.
}
\label{f-polygonal}
\end{figure*}

Fig.~\ref{f-polygonal} continues our running example,
and illustrates the construction of the polygonal curve.
The initial position $\pinit$ and terminal position $\pterm$ are shown
as black disks in the bottom left and bottom right, respectively.
The left figure shows the augmented line graph, where
these two points are connected to their adjacent vertices.
The initial point $\pinit$ has two adjacent vertices, while the
terminal point $\pterm$ has only one.
The center left figure shows the shortest path from the initial point
to the terminal point.
In the center right figure, we fix the boxes that
the curve must traverse, and we minimize the curve length by solving 
the SOCP~\eqref{e-shortening}.
In the right figure, a new box is inserted in the box sequence and the
curve nodes are optimized again.
In this example the polygonal phase converges in a single
iteration.

\section{Smooth Phase} \label{s-smooth}

The smooth phase is the final phase of our algorithm.
It starts from the polygonal curve $\mathcal C$
and constructs a smooth path $p$ that
is feasible for our planning problem, and has small objective value.
For the path parameterization we use
a piecewise B\'ezier curve, \ie, a sequence of B\'ezier
curves that connect smoothly.
(Sometimes this is also called a composite  B\'ezier curve.)
We start this section by reviewing some basic properties of this
family of curves.
Next we describe the optimal-control problems that we solve
to design our smooth path.
Finally, we conclude our running example.

\subsection{B\'ezier curves}

\begin{figure}
\centering
\includegraphics[width=.25\textwidth]{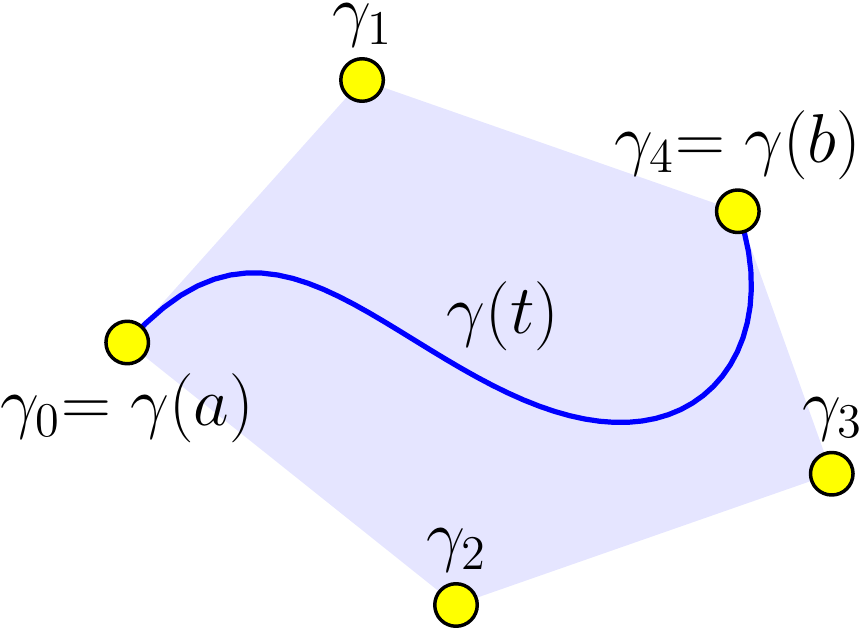}
\caption{B\'ezier curve with control points $\gamma_0,\ldots, \gamma_M$, $M=4$.
The curve starts at $\gamma(a) = \gamma_0$, ends at $\gamma(b) =\gamma_M$,
and is entirely contained  in the convex hull of the control points, shown shaded.}
\label{f-bez}
\end{figure}

A B\'ezier curve is constructed using Bernstein polynomials.
The Bernstein polynomials of degree $M$ are defined over
the interval $[a, b] \subset \reals$, with $b>a$, as
\[
\beta_n (t) = \binom{M}{n}
\left(\frac{t - a}{b - a}\right)^n
\left(\frac{b - t}{b - a}\right)^{M-n},
\quad n =0, \ldots, M.
\]
For $t \in [a,b]$ the Bernstein polynomials are nonnegative
and, by the binomial theorem, they sum up to one.
Therefore, the scalars $\beta_0(t), \ldots, \beta_M(t)$
can be thought of as the coefficients of a convex combination.
Using these coefficients to combine
a given set of \emph{control points} $\gamma_0, \ldots, \gamma_M \in \reals^d$,
we obtain a B\'ezier curve:
\[
\gamma(t) = \sum_{n = 0}^M \beta_n (t)  \gamma_n.
\]
The B\'ezier curve $\gamma:[a,b] \rightarrow \reals^d$ is a polynomial function of degree $M$.
An example of a B\'ezier curve is shown in Fig.~\ref{f-bez},
for $d=2$ and $M=4$.
Below we list some important properties that we will
use later in this section.

\paragraph*{Endpoints}
A B\'ezier curve starts at its first control
point and ends at its last control point, \ie,
\BEQ \label{e-ep-prop}
\gamma(a) = \gamma_0, \qquad \gamma(b) = \gamma_M.
\EEQ
With this property, a piecewise B\'ezier curve (our path)
can be made continuous simply by equating
the last control point of each curve piece with the first
control point of the next piece.

\paragraph*{Control polytope}
Since each point on a B\'ezier curve is a convex combination of the control
points, the entire curve is contained in the convex hull of the control points:
\BEQ \label{e-ch-prop}
\gamma(t) \in \conv \{\gamma_0, \ldots, \gamma_M\},
\EEQ
for all $t \in [a,b]$.
This convex hull is called the \emph{control polytope}
of the B\'ezier curve $\gamma$.
From this property it follows that if all control points lie in a convex
set (in our case a box), then so does the B\'ezier curve.

\paragraph*{Derivatives}
The derivative $\gamma^{(1)}$ of a B\'ezier curve $\gamma$ is
also a B\'ezier curve.
It has degree $M-1$ and its control points are
given by the difference equation
\BEQ \label{e-der-prop}
\gamma_n^{(1)} = \frac{M}{b-a} (\gamma_{n+1} - \gamma_n), \quad n=0,\ldots, M-1.
\EEQ
Iterating this, we see that the derivative $\gamma^{(i)}$ of any order
$i \geq 1$ is a B\'ezier curve of degree $M-i$.
Moreover, the derivatives of a piecewise B\'ezier curve
are also piecewise B\'ezier curves, and their continuity can be enforced
using the endpoint property~\eqref{e-ep-prop}.

\paragraph*{Squared $L_2$ norm}
The square of the $L_2$ norm of a B\'ezier curve $\gamma$
can be expressed as a function of the 
control points using the following expression~\cite[\S3.3]{farouki1988algorithms}:
\BEQ \label{e-integral-prop}
\int_a^b \|\gamma(t)\|_2^2 \; d t =
(b-a) Q (\gamma_0, \ldots, \gamma_M),
\EEQ
where $Q$ is a convex quadratic function defined as
\[
Q (\gamma_0, \ldots, \gamma_M) = 
\frac{1}{2M+1} \sum_{m=0}^M \sum_{n=0}^M
\frac{\binom{M}{m}\binom{M}{n}}{\binom{2M}{m + n}}
\gamma_m^T \gamma_n.
\]
This formula allows us also to compute the squared $L_2$ norm
of a piecewise B\'ezier curve and its derivatives.

\subsection{Nonconvex optimal-control problem}
\label{s-nonconvex}
In the smooth phase we limit our attention to paths
that are piecewise B\'ezier curves and traverse the
same box sequence $s_1, \ldots, s_N$ as the curve $\mathcal C$.
This reduces the path-planning problem~\eqref{e-ppp2} to an
optimal-control problem that is finite dimensional and has only
continuous variables, but is nonconvex.
This subsection illustrates this control problem,
and the next subsection describes our approach to its solution.

\paragraph*{Variables}
The variables in problem~\eqref{e-ppp2}
are the safety map $s$ and the path $p$.
Here the box sequence is fixed, therefore a safety map is specified
only through the traversal times $T_1, \ldots, T_N$, which are the
first variables in our control problem.
For the path parameterization we use a piecewise B\'ezier curve with $N$
pieces (one per safe box that our path must traverse).
Each piece, or subpath, is a B\'ezier curve
\[
p_j : [t_{j-1}, t_j] \rightarrow \reals^d, \qquad j=1, \ldots, N.
\]
These B\'ezier curves have degree equal to $M$, and their
control points,
\BEQ
\label{e-subpath-points}
p_{j,0}, \ldots, p_{j,M} \in \reals^d, \quad j=1, \ldots, N,
\EEQ
are the second group of variables in our control problem.

For $i=1,\ldots, D$,  the derivatives $p_j^{(i)}$ of the subpaths
are B\'ezier curves of degree $M-i$.
Our last set of variables are the control points
these derivatives:
\BEQ
\label{e-subpath-der-points}
p_{j,0}^{(i)}, \ldots, p_{j,M-i}^{(i)} \in \reals^d,  \quad i=1, \ldots, D,
\ j=1, \ldots, N.
\EEQ
For simplicity of notation, we will sometimes denote the control points
in~\eqref{e-subpath-points} as $p_{j,0}^{(0)}, \ldots, p_{j,M}^{(0)}$,
where the superscript represents the zeroth derivative.

\paragraph*{Constraints}
We assemble the constraints of our control problem
by leveraging the properties of the B\'ezier curves.

Using the endpoint property~\eqref{e-ep-prop},
the boundary conditions in problem~\eqref{e-ppp2}
are enforced simply as
\BEQ \label{e-bez-boundary}
p_{1,0} = \pinit, \qquad p_{N,M} = \pterm.
\EEQ
Similarly, the continuity and differentiability of our path
are enforced as
\BEQ \label{e-bez-continuity}
p_{j,M-i}^{(i)} = p_{j+1,0}^{(i)},
\quad i=0, \ldots, D,
\ j=1, \ldots, N-1.
\EEQ

Property~\eqref{e-ch-prop} tells us that a B\'ezier curve lies within its
control polytope.
Therefore, to ensure that a subpath $p_j$ is entirely contained in the
corresponding safe box $\mathcal B_{s_j}$,
it is sufficient to constrain its control points:
\BEQ \label{e-bez-safety}
l_{s_j} \leq p_{j,n} \leq u_{s_j},
\quad j=1, \ldots, N,
\ n=0, \ldots, M.
\EEQ
Since each subpath $p_j$ is constrained in a safe box,
the whole piecewise B\'ezier curve $p$ will be safe.

The control points of the subpath derivatives need to 
satisfy a difference equation analogous to~\eqref{e-der-prop}:
\begin{multline}
\label{e-bez-dynamics}
p_{j,n}^{(i)} = \frac{M - i + 1}{T_j}
\left(p_{j,n+1}^{(i-1)} - p_{j,n}^{(i-1)}\right), \\
\quad i=1, \ldots, D,
\ j=1, \ldots, N,
\ n = 0, \ldots, M - i.
\end{multline}
Note that these equality constraints are nonlinear,
since both the control points and the traversal times
are variables in our optimization problem.

Lastly, the traversal times need to be positive,
\BEQ \label{e-increasing}
T_j > 0, \qquad j=1, \ldots, N,
\EEQ
and sum up to the final time,
\BEQ \label{e-final-time}
\sum_{j=1}^N T_j = T.
\EEQ

\paragraph*{Objective function}
We split the integrals in our objective function~\eqref{e-obj}
into the sum of $N$ terms (one per subpath):
\[
J = \sum_{i=1}^D \alpha_i \sum_{j=1}^N J_{i,j}.
\]
We use~\eqref{e-integral-prop} to
express each term as a function of the control points
and the traversal times:
\BEQ
\label{e-subpath-cost}
J_{i,j} = \int_{t_{j-1}}^{t_j} \| p_j^{(i)} (t) \|_2^2 \; dt
= T_j Q(p_{j,0}^{(i)}, \ldots, p_{j,M-i}^{(i)}),
\EEQ
for $i=1, \ldots, D$ and $j=1, \ldots, N$.
Note that the quadratic function $Q$ is convex, but its product with
the traversal time $T_j$ makes our objective nonconvex.

\paragraph*{Optimization problem}
Collecting all the components, we obtain
the optimization problem
\BEQ \label{e-bez}
\begin{array}{ll}
\mbox{minimize} & J \\
\mbox{subject to}
& \text{constraints~\eqref{e-bez-boundary} to~\eqref{e-final-time}.}
\end{array}
\EEQ
This program has the structure of an optimal-control problem
where the difference equation~\eqref{e-bez-dynamics} acts as a
dynamical system that links the variables over time.
Together with the nonconvex objective terms~\eqref{e-subpath-cost},
this nonlinear difference equation makes the problem nonconvex.
However, similarly to its infinite-dimensional counterpart~\eqref{e-ppp2},
this problem simplifies to a convex quadratic program
(QP)~\cite[\S4.4]{boyd2004convex} if we fix the traversal times.
In fact, this makes the difference equation linear and the objective function
convex quadratic.

\paragraph*{Curve degree and feasibility}
If the degree of the subpaths satisfies
$M \geq 2D + 1$, then problem~\eqref{e-bez} is
guaranteed to be feasible.
In fact, similarly to the discussion in \S\ref{s-feas},
this minimum degree ensures that each subpath can be a line
segment, with the first $D$ derivatives equal to zero at the endpoints.
The overall path $p$ can then take the shape of the safe polygonal curve
$\mathcal C$, while satisfying the differentiability constraints.
Note also that the degree $M$ must be at least $D+1$, since
the continuity and differentiability constraints~\eqref{e-bez-continuity}
fix the value of $D+1$ control points per subpath.
In the rest of this paper we will use curves of degree $M=2D + 1$,
so that problem~\eqref{e-bez} will always be feasible.
Although, in practice, we have found that also values of $M$ closer to
$D+1$ almost always yield feasible problems.

\subsection{Solution via convex alternations}
\label{s-alternations}
We solve the nonconvex program~\eqref{e-bez} by alternating between
a \emph{projection problem} and a \emph{tangent problem},
both of which are convex optimal-control problems.
As the other parts of our planning method, this step is heuristic:
it is guaranteed to find a feasible solution of~\eqref{e-bez},
but this solution needs not to be optimal.

\paragraph*{Initialization}
We start by computing an initial estimate of
the traversal times $T_1, \ldots, T_N$ that satisfies
the constraints~\eqref{e-increasing} and~\eqref{e-final-time}.
To do so, we imagine  travelling along
the polygonal curve $\mathcal C$ at constant speed.
The window of time $T_j$
that we allocate for the $j$th box is then equal
to the distance between the nodes $y_j$ and $y_{j-1}$,
divided by the total length of the curve $\mathcal C$
and multiplied by the final time $T$.

Although this heuristic is very simple, we have found that it
works well for most problems.
More precise initial estimates of the traversal times are certainly
possible but, in our experience, they are rarely worth
their increased complexity.

\paragraph*{Projection problem}
In this step we fix the current value of the traversal times
(initialized as just described)
and we solve the control problem~\eqref{e-bez}.
As observed above this is a convex QP, which
has the effect of projecting the current iterate onto the
nonconvex feasible set of~\eqref{e-bez}.
Thanks to their optimal-control structure and their banded constraints,
these QPs are solvable in a time
that grows only linearly with the number $N$
of boxes traversed by our path~\cite{wang2009fast}.

\paragraph*{Tangent problem}
In this step we attempt to improve the estimate of the
traversal times by solving a convex approximation of~\eqref{e-bez}.

Let us introduce auxiliary variables that represent the products of
the traversal times and the control points of the path derivatives:
\BEQ \label{e-aux}
q_{j,n}^{(i)} = T_j p_{j,n}^{(i)},
\EEQ
for $i=1, \ldots, D$, $j=1, \ldots, N$, and $n = 0, \ldots, M - i$.
Using these variables, the nonlinear difference
equation~\eqref{e-bez-dynamics} becomes linear,
\[
q_{j,n}^{(i)} = (M - i + 1)
\left(p_{j,n+1}^{(i-1)} - p_{j,n}^{(i-1)}\right),
\]
and the nonconvex objective terms~\eqref{e-subpath-cost}
become quadratic over
linear~\cite[\S3.2.6]{boyd2004convex},
\[
J_{i,j} =
\frac{Q(q_{j,0}^{(i)}, \ldots, q_{j,M-i}^{(i)})}{T_j}.
\]
(Recall that quadratic-over-linear functions, with
the numerator convex and the denominator positive, are convex and
representable through a second-order
cone~\cite[\S2.3]{lobo1998applications}.)

The only nonconvexity left in our problem is
the nonlinear equality constraint~\eqref{e-aux},
which we simply linearize around the current
traversal times $\bar T_j$ and control points $\bar p_{j,n}^{(i)}$
(obtained by solving the projection problem):
\[
q_{j,n}^{(i)} =
- \bar T_j \bar p_{j,n}^{(i)}
+ T_j \bar p_{j,n}^{(i)}
+ \bar T_j p_{j,n}^{(i)}.
\]
Since this linearization might be inaccurate
away from the nominal point,
we also add a trust-region constraint
\BEQ \label{e-trust-region}
\frac{1}{1 + \kappa} \leq \frac{T_j}{\bar T_j}  \leq  1 + \kappa,
\qquad j=1, \ldots, N.
\EEQ
This sets a limit of $\kappa > 0$ to the maximum relative variation
of the traversal times.

The resulting problem is an SOCP that approximates
the nonconvex program~\eqref{e-bez} locally, and tries to
improve the current solution by taking a step in the
tangent space of the nonlinear equation~\eqref{e-aux}.
Like the projection problem, it
can be solved in a time that increases only linearly with $N$.
From its solution we only retain the optimal traversal times
$T_1^\star, \ldots, T_N^\star$, and then we solve a new projection problem
to obtain a new feasible path.
If the optimal objective value decreases, compared to the previous
projection problem, we accept the new times and update our path.
Otherwise we keep the previous times and path.

\paragraph*{Trust region update}
After each iteration, independently of its success,
we decrease the value of the trust-region parameter $\kappa$.
A simple way to do so would be to divide
$\kappa$ by a parameter $\omega>1$.
However, using this rule, we might have that
two consecutive iterations produce the same unsuccessful
update of the traversal times.
Specifically, if one iteration is unsuccessful and the
transition times computed in the tangent problem
are not at the boundary of the trust region~\eqref{e-trust-region},
then these times might still be feasible (and thus optimal)
after we shrink the trust region.
To prevent this phenomenon, we use
a slightly more sophisticated update:
\[
\kappa^+ = \frac{1}{\omega}
\left(
\max \left\{
\frac{\bar T_1}{T_1^\star}, \frac{T_1^\star}{\bar T_1},
\ldots,
\frac{\bar T_N}{T_N^\star}, \frac{T_N^\star}{\bar T_N}
\right\} - 1\right)
\leq \frac{\kappa}{\omega}.
\]
Here $\kappa$ and $\kappa^+$ are the current and the updated
trust-region parameters, respectively.
The term in the parenthesis is the minimum value of
$\kappa$ that, in hindsight, would have
activated at least one of the trust-region constraints~\eqref{e-trust-region}.
If one of these constraints was already active, then this
rule reduces to $\kappa^+ = \kappa / \omega$.
If none of the trust-region constraints was active,
and the iteration was unsuccessful,
then the trust region is shrunk enough to make the
solution of the last tangent problem infeasible for the next.

\paragraph*{Termination}
The tangent problem has optimal value smaller than or equal to
its preceding projection problem.
We terminate our algorithm when this gap, normalized by
the cost of the projection problem, is smaller than a
fixed tolerance $\varepsilon > 0$.
In which case, we solve one last projection problem and
we return the best path that we have found.

\paragraph*{Choice of the parameters}
We have found that for most problems the value of $\kappa$ can
be simply initialized to one.
Large values of $\omega$ (\eg, $\omega=5$) tend to work well when
our initialization of the traversal times is accurate, while smaller
values (\eg, $\omega=2$) are more effective otherwise.
In the numerical experiments discussed in this paper we
use $\omega=3$.
For the termination tolerance a reasonable choice is
$\varepsilon = 10^{-2}$.

\subsection{Example}

\begin{figure*}
\centering
\includegraphics[width=.23\textwidth]{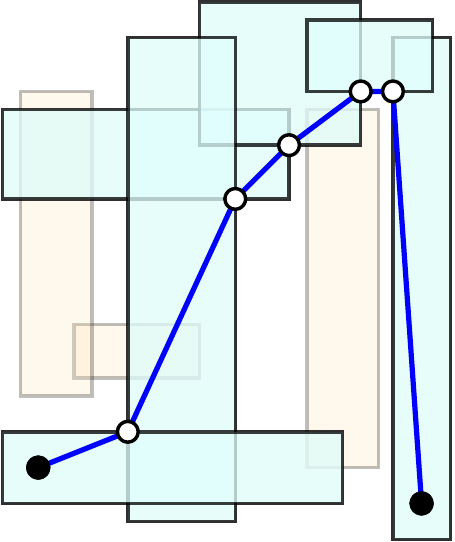} \quad
\includegraphics[width=.23\textwidth]{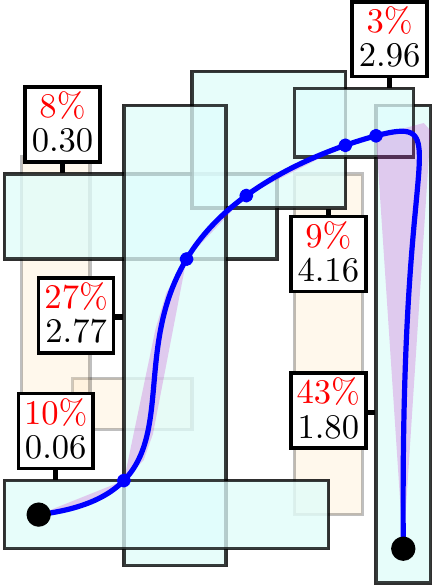} \quad
\includegraphics[width=.23\textwidth]{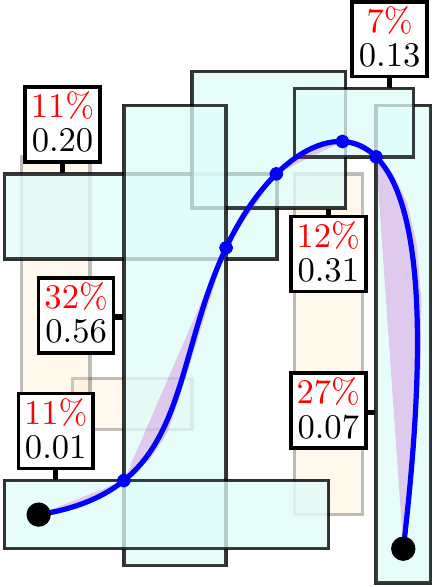} \quad
\includegraphics[width=.23\textwidth]{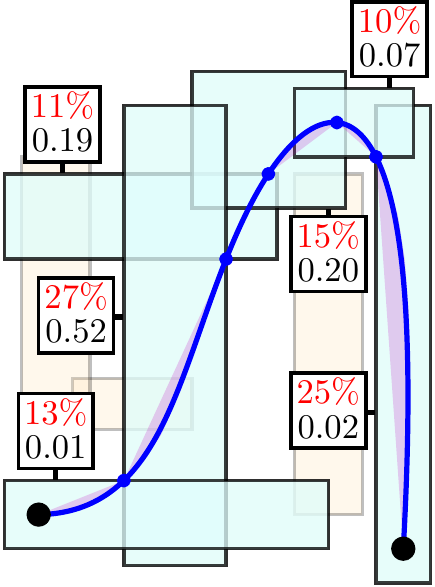}
\caption{Smooth phase of our algorithm.
\emph{Left.}
The curve from the polygonal phase
with the corresponding sequence of safe boxes.
\emph{Center left.}
The polygonal curve is used to estimate initial traversal
times and a first smooth path is optimized.
For each box traversed by the path,
the labels show the traversal time (normalized by the final time)
at the top, and the cost of the trajectory piece at the bottom.
The red shaded sets are the control polytopes of the B\'ezier curves.
\emph{Center right.}
The traversal times are improved
and the path is optimized a second time.
\emph{Right.}
The path at the last (fourth) iteration, whose cost
is within $0.01\%$ of the global minimum.}
\label{f-smooth}
\end{figure*}

We conclude our running example by illustrating the smooth phase of the
path-planning algorithm.
We seek a path that is $D=3$ times continuously differentiable,
and has total duration $T$ equal to one.
We use B\'ezier curves of degree $M=2D+1=7$.
We take objective weights $\alpha_1=\alpha_2=0$ and $\alpha_3>0$,
\ie, our objective penalizes the squared $L_2$ norm of the third derivative
(or jerk) of the path.
To simplify the analysis,  the weight $\alpha_3$ is chosen so
that the global minimum of the problem is equal to one.

In the left of Fig.~\ref{f-smooth} we have the curve computed in the polygonal phase,
with the corresponding sequence of safe boxes highlighted in cyan.
In the center left of Fig.~\ref{f-smooth} we show the path obtained
by solving the first projection problem, with the traversal times
initialized using the constant-velocity heuristic.
Each box traversed by the path is labeled with
two numbers: the percentage on top is the ratio between the
traversal time $T_j$ of that box and the final
time $T$; the number at the bottom is the cost $J_{3,j}$ of the
subpath $p_j$.
The red shaded areas are the B\'ezier control polytopes within each box.
We solve the tangent problem, we update the traversal times,
and we solve the projection problem a second time.
The resulting path is depicted in the center right of Fig.~\ref{f-smooth}.
After four iterations the smooth phase terminates, with the resulting
path depicted in the right of Fig.~\ref{f-smooth}.

The initial path (center left) has cost $12.04$.
The path after the first iteration (center right) has cost $1.27$, which is
$89\%$ smaller.
The final path (right) has cost $1.0001$, and
is essentially the global minimum of the problem.
Although our simple heuristic to initialize the
traversal times was not accurate, our algorithm converges
in very few iterations.

\section{Algorithm Efficiency and Guarantees}
\label{s-analysis}

We briefly summarize the main properties of our
path-planning method.

\paragraph*{Completeness}

Our algorithm is complete: it finds a safe smooth path
connecting the initial and final positions if
such a path exists, and it certifies infeasibility otherwise.
Feasibility is decided almost immediately with
the solutions of the shortest-path problem at the beginning of the polygonal phase.
If this problem is infeasible, then the initial and terminal points cannot
be connected by a continuous curve.
Conversely, if the shortest-path problem is feasible,
then our algorithm can always recover a smooth feasible path
as described at the end of \S\ref{s-nonconvex}.
Of course, if the safe boxes approximate
a more complex space,
then the completeness of our method is up to the
conservatism of this approximation.

\paragraph*{Suboptimality}

Our algorithm is heuristic, and not guaranteed to solve
problem~\eqref{e-ppp} optimally
(or within a fixed optimality tolerance).
In practice, the main source of suboptimality is the
choice of the box sequence, which is frozen after the polygonal
phase, and does not take into account the actual objective of
our path-planning problem.
Solving the nonconvex problem~\eqref{e-bez} is another
source of suboptimality, and our alternating method in
\S\ref{s-alternations} is designed to prioritize a low number of iterations
over the cost of the final path.
Two other (milder) approximations are the path
parameterization using B\'ezier curves, and the sufficiency
of the safety constraint~\eqref{e-bez-safety}.

Some heuristic steps in our path planner do not contribute
to the algorithm completeness, but can play an important role
in limiting the optimality losses just described.
For example, we could take the centers of the box intersections
as representative points, instead of optimizing them
as in~\eqref{e-rep-pts-opt}.
However, the downstream shortest-path problem
would typically select less efficient box sequences with this choice.
Two other main heuristic components of our method
are the iterative shortening of the polygonal curve in \S\ref{s-shortening} and
the initialization of the traversal times in \S\ref{s-alternations}.

\paragraph*{Runtimes}

The offline and online runtimes of our method
are dominated by the convex optimization problems.
The SOCP~\eqref{e-rep-pts-opt} is the preprocessing 
step that takes most time, but is efficiently solvable
even for path-planning problems of very large scale.
In addition, we note that this subproblem
only needs to be solved to modest, or even low, accuracy.
The SOCP~\eqref{e-shortening} in the polygonal phase
takes negligible time, since it is very small and
has banded constraints.
Furthermore, this phase usually converges
within four or five iterations.
(More formally, the iterations of this phase can be
bounded by the number $K$ of safe boxes;
since each iteration adds at least
one box to our sequence, and 
box repetitions are not optimal).
The projection QPs and the tangent SOCPs
in the smooth phase can be large problems,
but they take a time that is only linear in the number of traversed
boxes and can be solved to modest precision.
Note also that the trust region~\eqref{e-trust-region}
shrinks geometrically during the iterations of
the smooth phase.
Therefore, after a handful of iterations (typically four to eight with the parameters given in \S\ref{s-alternations})
the projection and the tangent problems are 
essentially identical, and the smooth phase terminates.

\section{Numerical Experiments}
\label{s-experiments}

In this section we analyze the performance of our method
through multiple numerical experiments.
Every experiment was run using the default values in our 
software implementation \texttt{fastpathplanning}, which we briefly describe below.
The computations were carried out on a computer with
2.4 GHz 8-Core Intel Core i9 processor and 64 GB of RAM.

For code readability and fast prototyping, the current
version of \texttt{fastpathplanning} uses \texttt{CVXPY}~\cite{diamond2016cvxpy} to
construct the convex optimization problems and pass them
to the solver.
This introduces an overhead that for some problems
can be even a few times larger than the actual solver times.
Since by communicating
directly with the solver this overhead can be made negligible,
the time spent within \texttt{CVXPY} has been eliminated from
the runtimes reported in this paper.

\subsection{Software package}

The algorithm presented in this paper is implemented
in the open-source Python software package \texttt{fastpathplanning},
which is available at \texttt{\url{https://github.com/cvxgrp/fastpathplanning}}.
For the graph computations (\eg, the construction of the line graph)
we use \texttt{NetworkX 3.2}~\cite{networkx}.
For the solution of the shortest-path problem in the line graph
we use \texttt{scipy 1.11.3}~\cite{scipy2020}.
The convex optimization problems are specified using
\texttt{CVXPY 1.4.1}~\cite{diamond2016cvxpy}, and solved with the 
\texttt{Clarabel 0.6.0} solver~\cite{clarabel}.

The following is a basic example of the usage of \texttt{fastpathplanning}.

\begin{lstlisting}[language=Python]
import fastpathplanning as fpp

# offline preprocessing
L = ... # lower bounds of the safe boxes
U = ... # upper bounds of the safe boxes
S = fpp.SafeSet(L, U)

# online path planning
p_init = ... # initial point
p_term = ... # terminal point
T = 1 # final time
alpha = [1, 1, 5] # cost weights
p = fpp.plan(S, p_init, p_term, T, alpha)

# evaluate solution
t = 0.5 # sample time
p_t = p(t)
\end{lstlisting}

The matrices \texttt{L} and \texttt{U} contain the lower bound
$l_k$ and the upper bound $u_k$ of each safe box $\mathcal B_k$,
$k=1, \ldots, K$.
These have dimension $K \times d$, and are not explicitly defined
in the code above.
In line~6 they are used to instantiate
the safe set $\mathcal S$ (as the object \texttt{S}).
This line is where the offline preprocessing is done, \ie,
we construct the line graph and optimize the representative points.
In line~13 the function \texttt{plan} finds a smooth path \texttt{p},
given the safe set, initial and terminal points, final time, and objective
coefficients.
The number $D$ of continuous derivatives that our path will have
is equal to the length of the list \texttt{alpha}.
By default, the degree of the B\'ezier curves is set to $M = 2D+1$.
The path object \texttt{p} can be called like a function by passing a time
$t \in [0, T]$ as in line~17.
(It also contains other attributes such as the list of B\'ezier control points and
the safe boxes $s_1, \ldots, s_N$ that the path traverses.)

\subsection{Scaling study}
\label{s-scaling-study}

In our first example we consider path-planning problems in
$d=2$ dimensions, and analyze the performance of our
algorithm as a function of the number $K$ of safe boxes.

We generate an instance of problem~\eqref{e-ppp}
as follows.
We construct a square grid with $P^2$ points
with integer coordinates $\{1, \ldots, P\}^2$.
We let each point in this grid be the center of a safe box $\mathcal B_k$.
Each box elongates
either horizontally or vertically, with equal probability. 
The short and long sides of
a box are drawn uniformly at random from the intervals $[0, 0.5]$ and $[0, 2]$,
respectively.

\begin{figure}
\centering
\includegraphics[width=\columnwidth]{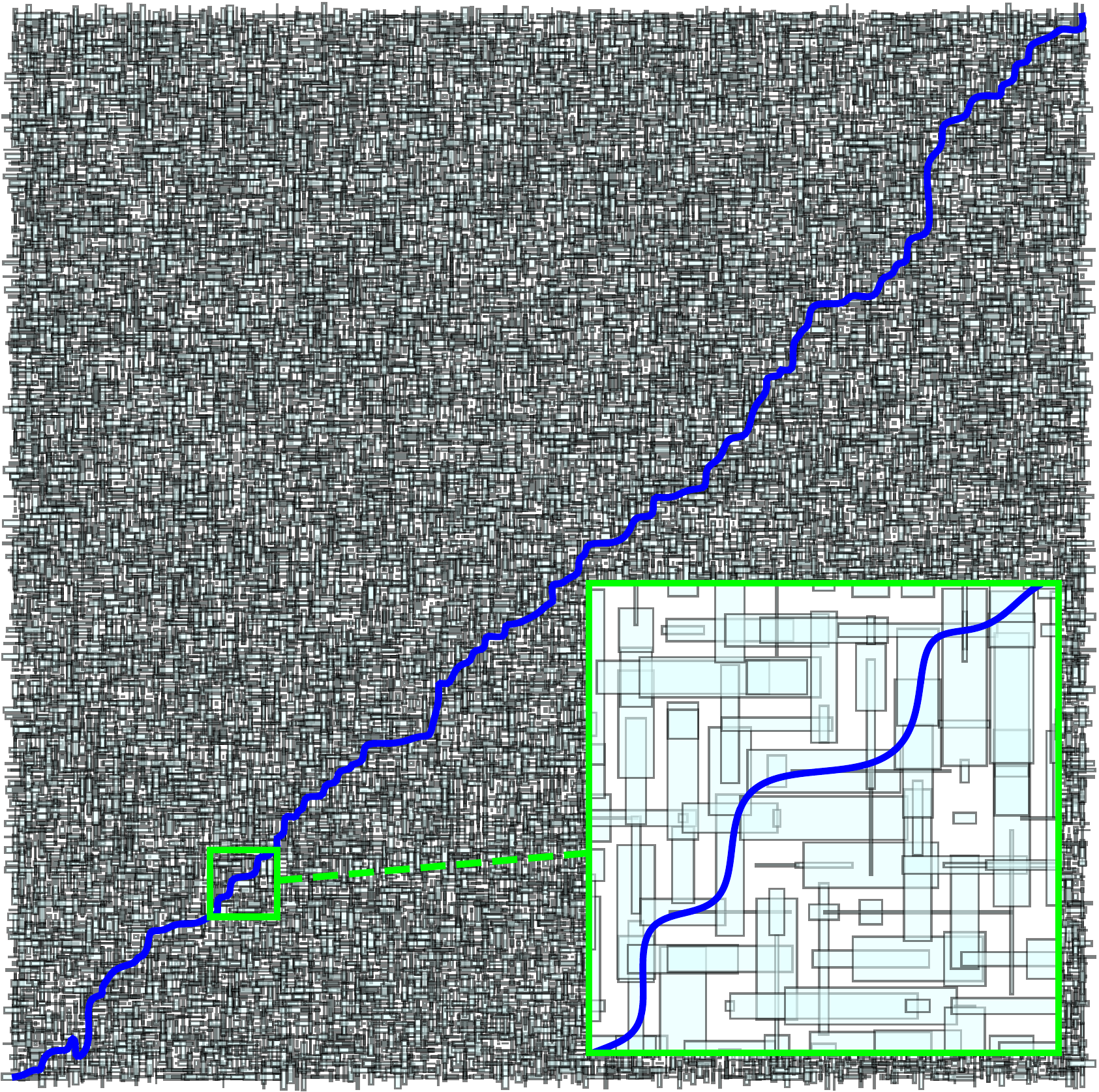}
\caption{Largest problem instance in the scaling study,
with $K=25{,}600$ safe boxes and final path shown.}
\label{f-smallest-largest}
\end{figure}

We use this procedure to generate six feasible path-planning
problems with grids of side $P=5,10,20,40,80,160$.
The number of boxes in these problems is then
\[
K = P^2 = 25, \ 100, \ 400, \ 1{,}600, \ 6{,}400, \ 25{,}600.
\]
The final time is taken to be $T=P$ and the cost weights
are $\alpha_1 = 0$ and $\alpha_2 = \alpha_3 = 1$.
The path is continuously differentiable $D=3$ times, and the
B\'ezier curves have degree $M = 2D+1 = 7$.
The initial position is  the center of the bottom-left box,
$\pinit = (1, 1)$, and the terminal position is
the center of the top-right box, $\pinit = (P, P)$.
The largest of these instances (with $K=25{,}600$ safe boxes) is
depicted in Fig.~\ref{f-smallest-largest}.

\begin{figure}
\centering
\includegraphics[width=\columnwidth]{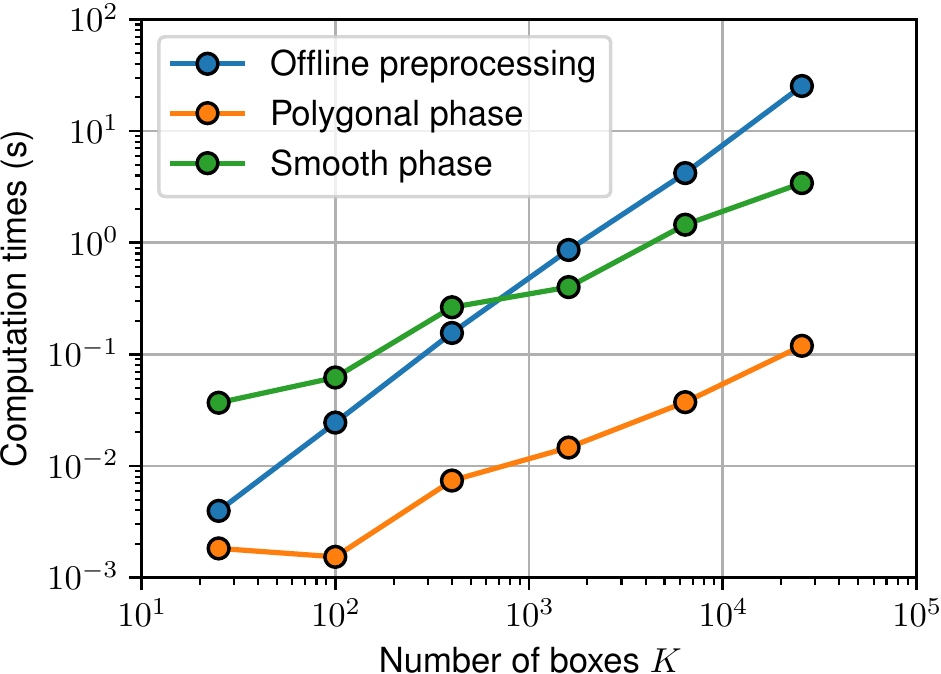}
\caption{Computation times for the scaling study, broken down into
offline preprocessing, polygonal phase, and smooth phase.
}
\label{f-comp-times}
\end{figure}

The computation times are shown in Fig.~\ref{f-comp-times},
broken down into offline preprocessing, polygonal phase, and smooth phase.
The smaller instances are solved in a few hundredths or tenths
of seconds.
For the largest instance in Fig.~\ref{f-smallest-largest}
the offline processing time is $25$ seconds,
while the online polygonal and smooth phases take $0.12$
and $3.4$ seconds, respectively.
Accounting for some fixed overhead, we see that the preprocessing
times grow almost linearly with the number of boxes $K$
(unit slope in the log-log plot), while the online runtimes
grow even more  slowly.
Tab.~\ref{t-scaling-study} shows the number of vertices
$|\mathcal V|$ and edges $|\mathcal E|$
in the line graph $G$ and the number of boxes $N$
traversed by the final path, for all the problems in this analysis.

We report that for both the  polygonal and
the smooth phase the number of iterations
is essentially unaffected by the size of the problem.
In the polygonal phase the number of iterations ranges between~1 and~4,
in the smooth phase between~5 and~6.

\begin{table}
\begin{center}
\caption{Size of the instances in the scaling study}
\begin{tabular}{ r|cccccc }
Total boxes $K$ & 25 & 100 & 400 & 1,600 & 6,400 & 25,600 \\ 
\hline
Vertices $|\mathcal V|$ & 38  & 179 & 804 & 3,298 & 12,816 & 52,308 \\ 
Edges $|\mathcal E|$ & 127 & 708 & 3,583 & 15,613 & 58,351 & 241,348 \\
Path boxes $N$ & 7 & 12 & 35 & 53 & 113 & 241
\end{tabular}
\label{t-scaling-study}
\end{center}
\end{table}

\subsection{Large example}
\label{s-quadrotor}

In our second example we plan a path
for a quadrotor in an environment with many obstacles.
The configuration space of a quadrotor is six dimensional:
three coordinates specify the position of the center of mass,
and three coordinates specify the orientation.
However, given any path for the center of mass
that is differentiable four times, a dynamically feasible
trajectory for the quadrotor's orientation,
together with the necessary control thrusts,
can always be reconstructed~\cite{mellinger2011minimum}.
This convenient property is called
\emph{differential flatness}, and it allows us to plan
the flight of a quadrotor by solving a path-planning problem in
only $d=3$ dimensions.

The quadrotor environment is shown at the top of Fig.~\ref{f-quad},
and it resembles a village with multiple buildings and dense vegetation.
This village is constructed over a square grid with
$P^2 = 50^2 = 2{,}500$ points,
which divide the ground into $(P-1)^2$ square cells of unit side.
The cell indexed by $(i,j) \in \{1, \ldots, P-1\}^2$ has bottom-left
coordinate $(i,j) \in \reals^2$ and top-right
coordinate $(i+1, j+1) \in \reals^2$.
Each cell contains one of the following obstacles:
a building, a bush, or a tree.
There are a total of $9^2 = 81$ buildings.
The cells that each building occupies are identified through a random
walk of length $5$ that starts in the cell with index
$(i,j) \in \{5, 10, \ldots, 40, 45\}^2$.
Therefore each building can cover up to six cells,
and neighboring buildings can potentially be connected.
The buildings are constructed so that the quadrotor,
whose collision geometry
is overestimated with a sphere of radius $0.1$, cannot
collide with them while flying in another cell.
The height of each building is equal to $5.0$.
In the cells that are not occupied by a building we have either a bush or
a tree, with equal probability.
Bushes and trees are positioned in the center of their cells.
A bush has square base of side chosen uniformly at random between
$0.2$ and $0.7$, and its height is twice the side of its base.
The foliage of a tree is represented as a cube of side $0.8$.
The center of the
foliage has height that is drawn uniformly at random between
$1.0$ and $4.5$.
The trunk of a tree has square section with side $0.2$.

To construct the safe set $\mathcal S$ we decompose the free space
in each cell independently using axis-aligned boxes.
The buildings occupy their cells entirely, so for these cells
we do not use any safe box.
The free space around a bush is decomposed using five safe boxes:
four around the bush and one on top.
Similarly, for a tree we have four safe boxes around the trunk and
one safe box on top of the foliage.
These boxes are appropriately shrunk to take into account the collision
geometry of the quadrotor.
The total number of safe boxes needed to decompose
the environment in Fig.~\ref{f-quad} using this method
is $K=10{,}150$.
The resulting line graph has $|\mathcal V| = 70{,}907$ vertices
and $|\mathcal E| = 1{,}022{,}782$ edges.

As shown in~\cite{mellinger2011minimum}, a natural objective
function when planning the path of a quadrotor is the squared $L_2$ norm
of the fourth derivative (or snap).
Thus we set our cost weights to
$\alpha_1 = \alpha_2 = \alpha_3 = 0$
and  $\alpha_4 = 1$.
We design a path that is continuously differentiable $D=4$ times,
and we use B\'ezier curves of degree $M = 2D+1 = 9$.
The final time is taken to be $T=P=50$.
The quadrotor takes off at the bottom left of the environment
$\pinit = (1, 1, 0)$, and lands in the top right
$\pinit = (P, P, 0)$.
Using the results from~\cite{mellinger2011minimum}, it can be seen that
for the quadrotor to start and stop horizontally,
with zero translational and angular velocity,
the following boundary conditions are necessary:
\[
p^{(i)} (0) = p^{(i)} (T) = 0, \quad i=1, \ldots, 3.
\]
The small modifications necessary for our algorithm to handle
these constraints are described in \S\ref{s-extensions}.

The offline preprocessing of the safe boxes takes $101$ seconds,
with the representative points
in~\eqref{e-rep-pts-opt} computed using the commercial solver
\texttt{MOSEK 10.0}.
The polygonal phase takes $0.22$ seconds, and it converges in
$5$ iterations.
The smooth phase takes $7.5$ seconds and $8$ iterations.
The number of boxes in the final path is $135$.
The bottom of Fig.~\ref{f-quad} shows the quadrotor flying
along the path generated by our algorithm.
A video of the quadrotor flight can be found at \texttt{\url{https://youtu.be/t9UWIi9NyxM}}.

In this example, as well as in any other problem where
we only penalize the path snap, our initial guess of the traversal times
is quite inaccurate, and the initial trajectory has very high cost.
However, the first iteration of the smooth phase is already
sufficient to reduce the cost by $84.3\%$, and the final trajectory
has a cost that is $99.3\%$ smaller than the initial one.

\subsection{Comparison with mixed-integer optimization}
\label{s-mip}

A very natural approach
to solving problem~\eqref{e-ppp} is mixed-integer (global)
optimization~\cite{deits2015efficient}.
We conclude our experiments with a comparison
of our method with these techniques.
As a benchmark we use our simple
running example illustrated in \S\ref{s-preprocessing}--\ref{s-smooth},
since the mixed-integer approach is impractical for larger problems.

To solve problem~\eqref{e-ppp} using mixed-integer optimization,
we parameterize a path as a piecewise B\'ezier curve with $N$
subpaths of equal duration $T_j = T/N$, $j=1, \ldots, N$.
We write a mixed-integer program that is identical to
problem~\eqref{e-bez}, except for the traversal times $T_j$ that here have
fixed value, and the
safety condition~\eqref{e-bez-safety} that is substituted with
a disjunctive constraint. This disjunctive constraint requires that each
subpath $p_j$ be contained in at least one safe box $\mathcal B_k$, and
is encoded using the binary variables $\sigma_{j,k} \in \{0,1\}$,
$j=1, \ldots, N$ and $k=1, \ldots, K$.
Since our safe sets are axis-aligned boxes, this constraint takes
the following simple form:
\[
\sum_{k=1}^K l_k \sigma_{j,k}
\leq p_{j,n}
\leq \sum_{k=1}^K u_k \sigma_{j,k},
\]
for $j=1, \ldots, N$ and $n=0, \ldots, M$.
The binary variables are also subject to the
``one-hot'' constraint
\[
\sum_{k=1}^K \sigma_{j,k} = 1,
\]
for $j=1, \ldots, N$.
The resulting problem is a mixed-integer quadratic program (MIQP).

We solve a sequence of MIQPs for an increasing number
of subpaths in our piecewise B\'ezier curve: $N=9, \ldots, 18$.
The minimum value $N=K=9$ is chosen since
the optimal path might have to visit each safe box.
Setting the degree of the B\'ezier curves to $M=2D+1=7$,
we then have that the mixed-integer approach features the
same completeness guarantee as our method, \ie, the MIQP is feasible if
and only if the original planning problem~\eqref{e-ppp} is feasible.
Larger values of $N$ yield a more flexible
path parameterization and can decrease the MIQP optimal value.
However, they also increase the MIQP solution times, which in the worst
case are proportional to the number $K^N$
of possible assignments of the binary variables.
Note that, since $N \geq K$, this worst-case runtime is super-exponential
in the number $K$ of boxes.

\begin{figure}
\centering
\includegraphics[width=\columnwidth]{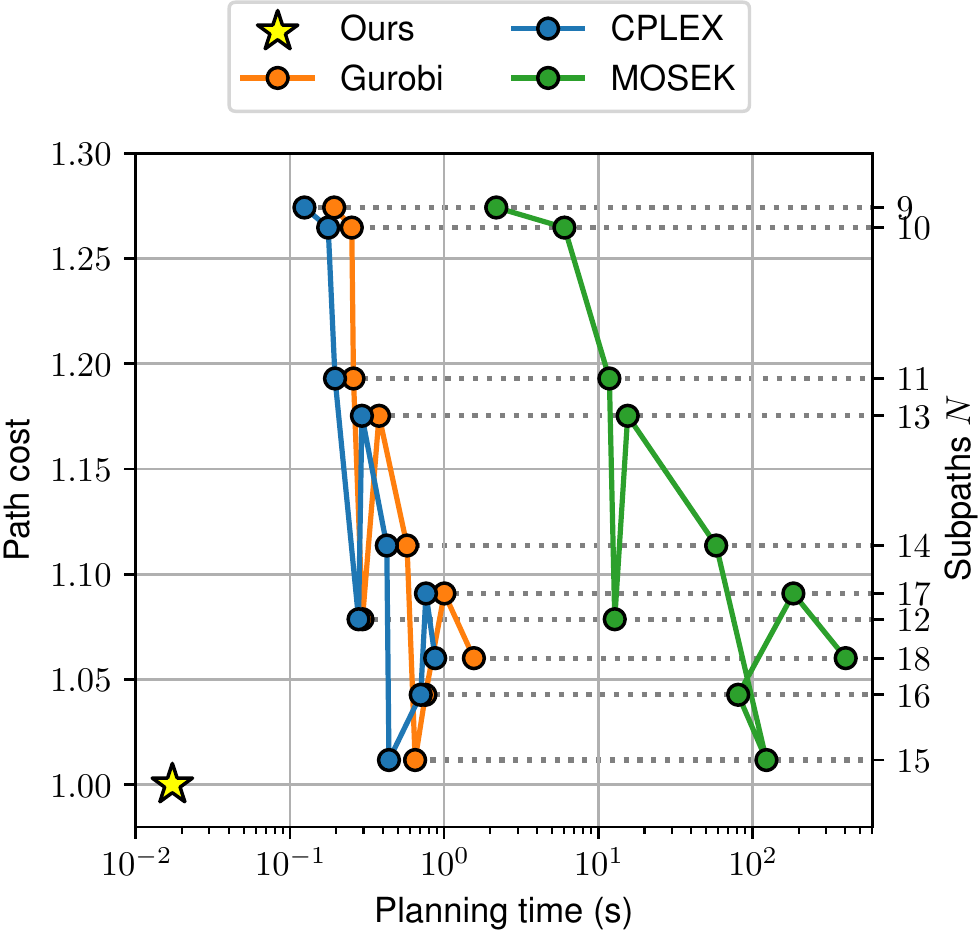}
\caption{Comparison of our algorithm with mixed-integer optimization.
The yellow star represents our method.
The three curves mark the performance of different commercial solvers
(\texttt{CPLEX}, \texttt{Gurobi}, and \texttt{MOSEK})
as the number $N$ of subpaths used
in the path parameterization increases.}
\label{f-mip}
\end{figure}

The path designed by our method for the running example
is illustrated in the right of Fig.~\ref{f-smooth},
has cost $1.0001$, and is essentially the global minimum of the problem
(which has unit cost).
The offline preprocessing (Fig.~\ref{f-line-graph-and-pts}),
the polygonal phase (Fig.~\ref{f-polygonal}), and the smooth
phase (Fig.~\ref{f-smooth}) of our algorithm take $1.0$, $1.7$, and $14.5$
milliseconds, respectively.
The sum of these three times ($17.2$ milliseconds)
and the cost of our path are reported in
Fig.~\ref{f-mip} with a yellow star.

For the solution of the MIQPs we consider three 
state-of-the-art commercial solvers:
\texttt{CPLEX 22.1.1}, \texttt{Gurobi 10.0}, and \texttt{MOSEK 10.0}.
Fig.~\ref{f-mip} reports the solution times and the path costs
for these solvers, as functions of the number $N$ of subpaths.
For $N=9$ the MIQP has an optimal value of $1.27$,
which is higher than our method since the duration of each subpath
is fixed in the MIQP, and the solver
cannot finely optimize the traversal times.
The fastest solver is \texttt{CPLEX} which takes
$124$ milliseconds, and is six times
slower than our approach.
The number of subpaths that leads to the MIQP of lowest optimal value is $N=15$.
This optimal value is $1.01$, which is closer to
but still larger than the cost of the path found with our method.
\texttt{CPLEX} is the fastest solver also in this case, and takes $439$ milliseconds
($26$ times slower than our algorithm).

As expected, the mixed-integer approach becomes quickly
impractical as the problem size grows.
For instance, by solving via MIQP the smallest problem in the scaling study
in \S\ref{s-scaling-study} (with $N=K$ and $M=2D+1$),
we get a path that is approximately three times cheaper than ours.
However, our planner takes only $43$ milliseconds
(including the offline preprocessing), while
\texttt{CPLEX} takes $584$ seconds to solve the MIQP,
and $25$ seconds of branch and bound before finding
a feasible path with cost lower than ours.
The other solvers are even slower.

\section{Extensions}
\label{s-extensions}

We conclude by briefly mentioning how the techniques
presented in this paper can be extended to
more general path-planning problems.

\paragraph*{Initial and final derivatives}
Our method handles boundary values on the path derivatives very easily.
We only need to specify them in problem~\eqref{e-bez} using the
control points $p_{1,0}^{(i)}$ and $p_{N,M-i}^{(i)}$,
with $i\in\{1,\ldots,D\}$, as done for the endpoint
constraints in~\eqref{e-bez-boundary}.

An additional modification to our algorithm
that is useful in presence of boundary conditions on the derivatives
concerns the estimate of the traversal times in \S\ref{s-alternations}.
Instead of initializing the traversal times by 
traveling the whole polygonal curve $\mathcal C$ at constant speed,
we travel at constant speed only the central part of $\mathcal C$,
and in the first and last segments we set to a constant the smallest
derivative that gives us enough variables to satisfy the boundary
conditions and the differentiability constraints.
For example, in the quadrotor problem in \S\ref{s-quadrotor},
we need constant seventh derivative in the
initial and final segments of $\mathcal C$ to
find a time parameterization whose first three derivatives
vanish at the endpoints, and that is continuously differentiable four times.

Finally, we note that with boundary conditions on the derivatives
the degree of the first and last B\'ezier curves might need to be increased
to preserve the completeness of our algorithm.
For example, if the initial position is close to a boundary
of the safe set, and the initial velocity points outwards,
we may need many control points to design a sharp turn that does not leave the safe set.

\paragraph*{Convex safe sets}
The assumption that the safe sets are axis-aligned boxes
is very convenient in the offline part of our algorithm,
since the pairwise intersections between a collection
of boxes can be found very efficiently~\cite{zomorodian2000fast}.
We also leveraged this assumption in the polygonal phase,
specifically in the multiple stabbing problems and in the 
improvement of the box sequence in \S\ref{s-shortening}.
In case of more generic convex safe sets these computations are
more demanding and can significantly slow down our algorithm.
For example, checking if two convex sets intersect
requires solving a convex optimization problem, \eg, a linear program 
when the sets are polyhedra.  However, if each convex
safe set is equipped with an axis-aligned bounding box,
part of the efficiency of our approach can be recovered.

\paragraph*{Unspecified final time}
In some applications specifying a fixed final time $T$ is not
straightforward, and it is preferable to let the planning algorithm
select this value automatically. In these cases, we also add a penalty
on $T$ (\eg, a linear cost $\alpha_0 T$ with fixed weight $\alpha_0 > 0$)
that prevents our original objective $J$ from making the 
final time arbitrarily large.
Our approach can be extended to these problems very naturally.
In the initialization of the
traversal times in \S\ref{s-alternations}, we now require an initial
guess also for the total duration of the path.
This guess is then improved by solving the
tangent problem in \S\ref{s-alternations},
where the final time $T$ is now a variable in the
linear constraint~\eqref{e-final-time},
and the cost function includes the time penalty
(\eg, $\alpha_0 T$).

\paragraph*{Derivative constraints}
Convex constraints on the path derivatives are also easily incorporated
in our framework.
In fact, the path derivatives are piecewise B\'ezier curves, and, similarly to
the safety constraints in~\eqref{e-bez-safety}, they can be forced
to lie in a convex set at all times by constraining their control points.
If the final time $T$ is fixed,
the addition of these constraints breaks the completeness of our algorithm.
Specifically, the feasibility argument in \S\ref{s-feas} does not hold anymore,
and the optimization of
our piecewise B\'ezier path in~\eqref{e-bez} might be infeasible
even if the original path-planning problem is feasible.
However, if we let $T$ be an optimization variable as described above,
then the algorithm completeness is recovered. This because any derivative
constraint (that contains the origin in its interior) can be satisfied by
travelling along a  curve sufficiently slowly.

\paragraph*{Multiple waypoints}
In some path-planning problems we need to design
a single smooth path that interpolates or passes through
a given sequence of intermediate waypoints in order.
To extend our approach to these problems, 
the steps in the polygonal phase are repeated to connect each
pair of consecutive waypoints, yielding a single polygonal
curve that satisfies all the interpolation constraints.
Similarly, in the smooth phase, we concatenate multiple problems
of the form~\eqref{e-bez} into a single program,
where each piecewise B\'ezier
curve has fixed endpoints and is constrained to connect smoothly with its
neighbors.
The time at which the overall path visits each waypoint is
then automatically selected by the smooth phase.
Finally, periodic trajectories that visit all the waypoints can be
generated by asking our path to satisfy
$p^{(i)} (0) = p^{(i)} (T)$, $i=0,\ldots,D$.
These conditions translate immediately to linear constraints on the control
points of the initial and final B\'ezier subpaths.

\section*{Acknowledgments}

This research was supported by the Office of Naval Research (ONR),
Award No. N00014-22-1-2121.
Indeed, this work is a direct consequence of the collaboration
fostered by this grant.
Parth Nobel was supported in part by the National Science Foundation
Graduate Research Fellowship Program under Grant No. DGE-1656518. Any
opinions, findings, and conclusions or recommendations expressed in
this material are those of the author(s) and do not necessarily
reflect the views of the National Science Foundation.  
Stephen Boyd was partially supported by ACCESS (AI Chip Center for Emerging
Smart Systems), sponsored by InnoHK funding, Hong Kong SAR.

\bibliographystyle{IEEEtran}
\bibliography{fpp.bib}

\begin{thebibliography}{10}
\providecommand{\url}[1]{#1}
\csname url@samestyle\endcsname
\providecommand{\newblock}{\relax}
\providecommand{\bibinfo}[2]{#2}
\providecommand{\BIBentrySTDinterwordspacing}{\spaceskip=0pt\relax}
\providecommand{\BIBentryALTinterwordstretchfactor}{4}
\providecommand{\BIBentryALTinterwordspacing}{\spaceskip=\fontdimen2\font plus
\BIBentryALTinterwordstretchfactor\fontdimen3\font minus
  \fontdimen4\font\relax}
\providecommand{\BIBforeignlanguage}[2]{{%
\expandafter\ifx\csname l@#1\endcsname\relax
\typeout{** WARNING: IEEEtran.bst: No hyphenation pattern has been}%
\typeout{** loaded for the language `#1'. Using the pattern for}%
\typeout{** the default language instead.}%
\else
\language=\csname l@#1\endcsname
\fi
#2}}
\providecommand{\BIBdecl}{\relax}
\BIBdecl

\bibitem{lavalle1997motion}
S.~LaValle and R.~Sharma, ``On motion planning in changing, partially
  predictable environments,'' \emph{The International Journal of Robotics
  Research}, vol.~16, no.~6, pp. 775--805, 1997.

\bibitem{fiorini1998motion}
P.~Fiorini and Z.~Shiller, ``Motion planning in dynamic environments using
  velocity obstacles,'' \emph{The international journal of robotics research},
  vol.~17, no.~7, pp. 760--772, 1998.

\bibitem{petti2005safe}
S.~Petti and T.~Fraichard, ``Safe motion planning in dynamic environments,'' in
  \emph{International Conference on Intelligent Robots and Systems}.\hskip 1em
  plus 0.5em minus 0.4em\relax IEEE/RJS, 2005, pp. 2210--2215.

\bibitem{du2011robot}
N.~Du~Toit and J.~Burdick, ``Robot motion planning in dynamic, uncertain
  environments,'' \emph{IEEE Transactions on Robotics}, vol.~28, no.~1, pp.
  101--115, 2011.

\bibitem{frazzoli2002real}
E.~Frazzoli, M.~Dahleh, and E.~Feron, ``Real-time motion planning for agile
  autonomous vehicles,'' \emph{Journal of guidance, control, and dynamics},
  vol.~25, no.~1, pp. 116--129, 2002.

\bibitem{kuwata2009real}
Y.~Kuwata, J.~Teo, G.~Fiore, S.~Karaman, E.~Frazzoli, and J.~P. How,
  ``Real-time motion planning with applications to autonomous urban driving,''
  \emph{IEEE Transactions on control systems technology}, vol.~17, no.~5, pp.
  1105--1118, 2009.

\bibitem{katrakazas2015real}
C.~Katrakazas, M.~Quddus, W.-H. Chen, and L.~Deka, ``Real-time motion planning
  methods for autonomous on-road driving: State-of-the-art and future research
  directions,'' \emph{Transportation Research Part C: Emerging Technologies},
  vol.~60, pp. 416--442, 2015.

\bibitem{sadigh2016planning}
D.~Sadigh, S.~Sastry, S.~A. Seshia, and A.~D. Dragan, ``Planning for autonomous
  cars that leverage effects on human actions,'' in \emph{Robotics: Science and
  systems}, vol.~2.\hskip 1em plus 0.5em minus 0.4em\relax Ann Arbor, MI, USA,
  2016, pp. 1--9.

\bibitem{wang2017safety}
L.~Wang, A.~D. Ames, and M.~Egerstedt, ``Safety barrier certificates for
  collisions-free multirobot systems,'' \emph{IEEE Transactions on Robotics},
  vol.~33, no.~3, pp. 661--674, 2017.

\bibitem{liniger2019noncooperative}
A.~Liniger and J.~Lygeros, ``A noncooperative game approach to autonomous
  racing,'' \emph{IEEE Transactions on Control Systems Technology}, vol.~28,
  no.~3, pp. 884--897, 2019.

\bibitem{spica2020real}
R.~Spica, E.~Cristofalo, Z.~Wang, E.~Montijano, and M.~Schwager, ``A real-time
  game theoretic planner for autonomous two-player drone racing,'' \emph{IEEE
  Transactions on Robotics}, vol.~36, no.~5, pp. 1389--1403, 2020.

\bibitem{deits2015efficient}
R.~Deits and R.~Tedrake, ``Efficient mixed-integer planning for {UAV}s in
  cluttered environments,'' in \emph{International Conference on Robotics and
  Automation}.\hskip 1em plus 0.5em minus 0.4em\relax IEEE, 2015, pp. 42--49.

\bibitem{marcucci2023motion}
T.~Marcucci, M.~Petersen, D.~von Wrangel, and R.~Tedrake, ``Motion planning
  around obstacles with convex optimization,'' \emph{Science Robotics}, vol.~8,
  no.~84, p. eadf7843, 2023.

\bibitem{lien2004approximate}
J.-M. Lien and N.~Amato, ``Approximate convex decomposition of polygons,'' in
  \emph{Proceedings of the twentieth annual symposium on Computational
  geometry}, 2004, pp. 17--26.

\bibitem{ayanian10}
N.~Ayanian and V.~Kumar, ``Abstractions and controllers for groups of robots in
  environments with obstacles,'' in \emph{International Conference on Robotics
  and Automation}.\hskip 1em plus 0.5em minus 0.4em\relax IEEE, 2010.

\bibitem{ghosh2013fast}
M.~Ghosh, N.~M. Amato, Y.~Lu, and J.-M. Lien, ``Fast approximate convex
  decomposition using relative concavity,'' \emph{Computer-Aided Design},
  vol.~45, no.~2, pp. 494--504, 2013.

\bibitem{deits2015computing}
R.~Deits and R.~Tedrake, ``Computing large convex regions of obstacle-free
  space through semidefinite programming,'' in \emph{Algorithmic Foundations of
  Robotics XI}.\hskip 1em plus 0.5em minus 0.4em\relax Springer, 2015, pp.
  109--124.

\bibitem{werner2023approximating}
P.~Werner, A.~Amice, T.~Marcucci, D.~Rus, and R.~Tedrake, ``Approximating robot
  configuration spaces with few convex sets using clique covers of visibility
  graphs,'' \emph{arXiv preprint arXiv:2310.02875}, 2023.

\bibitem{amice2022finding}
A.~Amice, H.~Dai, P.~Werner, A.~Zhang, and R.~Tedrake, ``Finding and optimizing
  certified, collision-free regions in configuration space for robot
  manipulators,'' in \emph{Algorithmic Foundations of Robotics XV}.\hskip 1em
  plus 0.5em minus 0.4em\relax Springer, 2022, pp. 328--348.

\bibitem{verghese2022configuration}
M.~Verghese, N.~Das, Y.~Zhi, and M.~Yip, ``Configuration space decomposition
  for scalable proxy collision checking in robot planning and control,''
  \emph{IEEE Robotics and Automation Letters}, vol.~7, no.~2, pp. 3811--3818,
  2022.

\bibitem{dai2023certified}
H.~Dai, A.~Amice, P.~Werner, A.~Zhang, and R.~Tedrake, ``Certified polyhedral
  decompositions of collision-free configuration space,'' \emph{arXiv preprint
  arXiv:2302.12219}, 2023.

\bibitem{petersen2023growing}
M.~Petersen and R.~Tedrake, ``Growing convex collision-free regions in
  configuration space using nonlinear programming,'' \emph{arXiv preprint
  arXiv:2303.14737}, 2023.

\bibitem{wang2009fast}
Y.~Wang and S.~Boyd, ``Fast model predictive control using online
  optimization,'' \emph{IEEE Transactions on control systems technology},
  vol.~18, no.~2, pp. 267--278, 2009.

\bibitem{lavalle2006planning}
S.~LaValle, \emph{Planning algorithms}.\hskip 1em plus 0.5em minus 0.4em\relax
  Cambridge university press, 2006.

\bibitem{marcucci2021shortest}
T.~Marcucci, J.~Umenberger, P.~Parrilo, and R.~Tedrake, ``Shortest paths in
  graphs of convex sets,'' \emph{arXiv preprint arXiv:2101.11565v4}, 2021.

\bibitem{schouwenaars2001mixed}
T.~Schouwenaars, B.~De~Moor, E.~Feron, and J.~How, ``Mixed integer programming
  for multi-vehicle path planning,'' in \emph{European control
  conference}.\hskip 1em plus 0.5em minus 0.4em\relax IEEE, 2001, pp.
  2603--2608.

\bibitem{richards2002coordination}
A.~Richards, J.~Bellingham, M.~Tillerson, and J.~How, ``Coordination and
  control of multiple uavs,'' in \emph{AIAA guidance, navigation, and control
  conference and exhibit}, 2002, p. 4588.

\bibitem{mellinger2012mixed}
D.~Mellinger, A.~Kushleyev, and V.~Kumar, ``Mixed-integer quadratic program
  trajectory generation for heterogeneous quadrotor teams,'' in
  \emph{International Conference on Robotics and Automation}.\hskip 1em plus
  0.5em minus 0.4em\relax IEEE, 2012, pp. 477--483.

\bibitem{augugliaro2012generation}
F.~Augugliaro, A.~Schoellig, and R.~D'Andrea, ``Generation of collision-free
  trajectories for a quadrocopter fleet: A sequential convex programming
  approach,'' in \emph{International Conference on Intelligent Robots and
  Systems}.\hskip 1em plus 0.5em minus 0.4em\relax IEEE/RJS, 2012, pp.
  1917--1922.

\bibitem{schulman2014motion}
J.~Schulman, Y.~Duan, J.~Ho, A.~Lee, I.~Awwal, H.~Bradlow, J.~Pan, S.~Patil,
  K.~Goldberg, and P.~Abbeel, ``Motion planning with sequential convex
  optimization and convex collision checking,'' \emph{The International Journal
  of Robotics Research}, vol.~33, no.~9, pp. 1251--1270, 2014.

\bibitem{liu2014solving}
X.~Liu and P.~Lu, ``Solving nonconvex optimal control problems by convex
  optimization,'' \emph{Journal of Guidance, Control, and Dynamics}, vol.~37,
  no.~3, pp. 750--765, 2014.

\bibitem{majumdar2017funnel}
A.~Majumdar and R.~Tedrake, ``Funnel libraries for real-time robust feedback
  motion planning,'' \emph{The International Journal of Robotics Research},
  vol.~36, no.~8, pp. 947--982, 2017.

\bibitem{bonalli2019gusto}
R.~Bonalli, A.~Cauligi, A.~Bylard, and M.~Pavone, ``Gusto: Guaranteed
  sequential trajectory optimization via sequential convex programming,'' in
  \emph{2019 International conference on robotics and automation (ICRA)}.\hskip
  1em plus 0.5em minus 0.4em\relax IEEE, 2019, pp. 6741--6747.

\bibitem{zhang2020optimization}
X.~Zhang, A.~Liniger, and F.~Borrelli, ``Optimization-based collision
  avoidance,'' \emph{IEEE Transactions on Control Systems Technology}, vol.~29,
  no.~3, pp. 972--983, 2020.

\bibitem{kavraki1996probabilistic}
L.~Kavraki, P.~Svestka, J.-C. Latombe, and M.~Overmars, ``Probabilistic
  roadmaps for path planning in high-dimensional configuration spaces,''
  \emph{IEEE Transactions on Robotics and Automation}, vol.~12, no.~4, pp.
  566--580, 1996.

\bibitem{lavalle1998rapidly}
S.~LaValle, ``Rapidly-exploring random trees: A new tool for path planning,''
  \emph{TR 98-11, Computer Science Department, Iowa State University}, 1998.

\bibitem{karaman2011sampling}
S.~Karaman and E.~Frazzoli, ``Sampling-based algorithms for optimal motion
  planning,'' \emph{The international journal of robotics research}, vol.~30,
  no.~7, pp. 846--894, 2011.

\bibitem{ratliff2009chomp}
N.~Ratliff, M.~Zucker, J.~A. Bagnell, and S.~Srinivasa, ``Chomp: Gradient
  optimization techniques for efficient motion planning,'' in \emph{2009 IEEE
  International Conference on Robotics and Automation}.\hskip 1em plus 0.5em
  minus 0.4em\relax IEEE, 2009, pp. 489--494.

\bibitem{kalakrishnan2011stomp}
M.~Kalakrishnan, S.~Chitta, E.~Theodorou, P.~Pastor, and S.~Schaal, ``{STOMP}:
  Stochastic trajectory optimization for motion planning,'' in \emph{2011 IEEE
  International Conference on Robotics and Automation}.\hskip 1em plus 0.5em
  minus 0.4em\relax IEEE, 2011, pp. 4569--4574.

\bibitem{hauser2016learning}
K.~Hauser, ``Learning the problem-optimum map: Analysis and application to
  global optimization in robotics,'' \emph{IEEE Transactions on Robotics},
  vol.~33, no.~1, pp. 141--152, 2016.

\bibitem{boyd2004convex}
S.~Boyd and L.~Vandenberghe, \emph{Convex Optimization}.\hskip 1em plus 0.5em
  minus 0.4em\relax Cambridge University Press, 2004.

\bibitem{belta2005discrete}
C.~Belta, V.~Isler, and G.~J. Pappas, ``Discrete abstractions for robot motion
  planning and control in polygonal environments,'' \emph{IEEE Transactions on
  Robotics}, vol.~21, no.~5, pp. 864--874, 2005.

\bibitem{alur2000discrete}
R.~Alur, T.~A. Henzinger, G.~Lafferriere, and G.~J. Pappas, ``Discrete
  abstractions of hybrid systems,'' \emph{Proceedings of the IEEE}, vol.~88,
  no.~7, pp. 971--984, 2000.

\bibitem{tedrake2010lqr}
R.~Tedrake, I.~Manchester, M.~Tobenkin, and J.~Roberts, ``Lqr-trees: Feedback
  motion planning via sums-of-squares verification,'' \emph{The International
  Journal of Robotics Research}, vol.~29, no.~8, pp. 1038--1052, 2010.

\bibitem{singh2017robust}
S.~Singh, A.~Majumdar, J.-J. Slotine, and M.~Pavone, ``Robust online motion
  planning via contraction theory and convex optimization,'' in \emph{2017 IEEE
  International Conference on Robotics and Automation (ICRA)}.\hskip 1em plus
  0.5em minus 0.4em\relax IEEE, 2017, pp. 5883--5890.

\bibitem{ames2016control}
A.~D. Ames, X.~Xu, J.~W. Grizzle, and P.~Tabuada, ``Control barrier function
  based quadratic programs for safety critical systems,'' \emph{IEEE
  Transactions on Automatic Control}, vol.~62, no.~8, pp. 3861--3876, 2016.

\bibitem{weiss2014spacecraft}
A.~Weiss, F.~Leve, M.~Baldwin, J.~R. Forbes, and I.~Kolmanovsky, ``Spacecraft
  constrained attitude control using positively invariant constraint admissible
  sets on {$SO(3) \times \mathbb R^3$},'' in \emph{2014 American Control
  Conference}.\hskip 1em plus 0.5em minus 0.4em\relax IEEE, 2014, pp.
  4955--4960.

\bibitem{weiss2015safe}
A.~Weiss, C.~Petersen, M.~Baldwin, R.~S. Erwin, and I.~Kolmanovsky, ``Safe
  positively invariant sets for spacecraft obstacle avoidance,'' \emph{Journal
  of Guidance, Control, and Dynamics}, vol.~38, no.~4, pp. 720--732, 2015.

\bibitem{berntorp2019positive}
K.~Berntorp, R.~Bai, K.~F. Erliksson, C.~Danielson, A.~Weiss, and
  S.~Di~Cairano, ``Positive invariant sets for safe integrated vehicle motion
  planning and control,'' \emph{IEEE Transactions on Intelligent Vehicles},
  vol.~5, no.~1, pp. 112--126, 2019.

\bibitem{danielson2020robust}
C.~Danielson, K.~Berntorp, A.~Weiss, and S.~Di~Cairano, ``Robust motion
  planning for uncertain systems with disturbances using the invariant-set
  motion planner,'' \emph{IEEE Transactions on Automatic Control}, vol.~65,
  no.~10, pp. 4456--4463, 2020.

\bibitem{choi2008path}
J.-w. Choi, R.~Curry, and G.~Elkaim, ``Path planning based on b{\'e}zier curve
  for autonomous ground vehicles,'' in \emph{Advances in Electrical and
  Electronics Engineering-IAENG Special Edition of the World Congress on
  Engineering and Computer Science 2008}.\hskip 1em plus 0.5em minus
  0.4em\relax IEEE, 2008, pp. 158--166.

\bibitem{flores2008real}
M.~Flores, \emph{Real-time trajectory generation for constrained nonlinear
  dynamical systems using non-uniform rational b-spline basis functions}.\hskip
  1em plus 0.5em minus 0.4em\relax California Institute of Technology, 2008.

\bibitem{lau2009kinodynamic}
B.~Lau, C.~Sprunk, and W.~Burgard, ``Kinodynamic motion planning for mobile
  robots using splines,'' in \emph{International Conference on Intelligent
  Robots and Systems}.\hskip 1em plus 0.5em minus 0.4em\relax IEEE/RJS, 2009,
  pp. 2427--2433.

\bibitem{elbanhawi2015continuous}
M.~Elbanhawi, M.~Simic, and R.~Jazar, ``Continuous path smoothing for car-like
  robots using b-spline curves,'' \emph{Journal of Intelligent \& Robotic
  Systems}, vol.~80, pp. 23--56, 2015.

\bibitem{park2019fast}
J.~Park and H.~J. Kim, ``Fast trajectory planning for multiple quadrotors using
  relative safe flight corridor,'' in \emph{International Conference on
  Intelligent Robots and Systems}.\hskip 1em plus 0.5em minus 0.4em\relax
  IEEE/RSJ, 2019, pp. 596--603.

\bibitem{koolen2020balance}
F.~Koolen, ``Balance control and locomotion planning for humanoid robots using
  nonlinear centroidal models,'' Ph.D. dissertation, Massachusetts Institute of
  Technology, 2020.

\bibitem{tordesillas2021faster}
J.~Tordesillas, B.~T. Lopez, M.~Everett, and J.~P. How, ``Faster: Fast and safe
  trajectory planner for navigation in unknown environments,'' \emph{IEEE
  Transactions on Robotics}, vol.~38, no.~2, pp. 922--938, 2021.

\bibitem{csomay2022multi}
N.~Csomay-Shanklin, A.~Taylor, U.~Rosolia, and A.~Ames, ``Multi-rate planning
  and control of uncertain nonlinear systems: Model predictive control and
  control {L}yapunov functions,'' \emph{arXiv preprint arXiv:2204.00152}, 2022.

\bibitem{arslan2022adaptive}
{\"O}.~Arslan and A.~Tiemessen, ``Adaptive b{\'e}zier degree reduction and
  splitting for computationally efficient motion planning,'' \emph{IEEE
  Transactions on Robotics}, vol.~38, no.~6, pp. 3655--3674, 2022.

\bibitem{zomorodian2000fast}
A.~Zomorodian and H.~Edelsbrunner, ``Fast software for box intersections,'' in
  \emph{Proceedings of the sixteenth annual symposium on computational
  geometry}, 2000, pp. 129--138.

\bibitem{lobo1998applications}
M.~Lobo, L.~Vandenberghe, S.~Boyd, and H.~Lebret, ``Applications of
  second-order cone programming,'' \emph{Linear algebra and its applications},
  vol. 284, no. 1-3, pp. 193--228, 1998.

\bibitem{scipy2020}
P.~Virtanen \emph{et~al.}, ``{{SciPy} 1.0: Fundamental Algorithms for
  Scientific Computing in Python},'' \emph{Nature Methods}, vol.~17, pp.
  261--272, 2020.

\bibitem{farouki1988algorithms}
R.~Farouki and V.~Rajan, ``Algorithms for polynomials in {B}ernstein form,''
  \emph{Computer Aided Geometric Design}, vol.~5, no.~1, pp. 1--26, 1988.

\bibitem{diamond2016cvxpy}
S.~Diamond and S.~Boyd, ``{CVXPY}: {A} {P}ython-embedded modeling language for
  convex optimization,'' \emph{Journal of Machine Learning Research}, vol.~17,
  no.~83, pp. 1--5, 2016.

\bibitem{networkx}
A.~Hagberg, D.~Schult, and P.~Swart, ``Exploring network structure, dynamics,
  and function using {N}etwork{X},'' in \emph{Proceedings of the 7th Python in
  Science Conference}, G.~Varoquaux, T.~Vaught, and J.~Millman, Eds., Pasadena,
  CA USA, 2008, pp. 11 -- 15.

\bibitem{clarabel}
\BIBentryALTinterwordspacing
P.~Goulart and Y.~Chen, ``Clarabel solver documentation,'' 2023. [Online].
  Available: \url{https://oxfordcontrol.github.io/ClarabelDocs/stable/}
\BIBentrySTDinterwordspacing

\bibitem{mellinger2011minimum}
D.~Mellinger and V.~Kumar, ``Minimum snap trajectory generation and control for
  quadrotors,'' in \emph{International Conference on Robotics and
  Automation}.\hskip 1em plus 0.5em minus 0.4em\relax IEEE, 2011, pp.
  2520--2525.

\end{thebibliography}

\appendices

\section{}
\label{s-box-update}

In this appendix we derive the inequalities~\eqref{e-box-update}, which are used
in the polygonal phase to improve the box sequence traversed by
the curve $\mathcal C$.

To simplify the notation, in this appendix we
let $a = y_{j-1}$, $b = y_{j+1}$, and $y = y_j$.
In addition, we denote with $l$ and $u$ the
lower and upper bounds that delimit the
axis-aligned box $\mathcal B_{s_j} \cap \mathcal B_{s_{j+1}}$.
Similarly, we let $l_1$ and $u_1$ delimit the box
$\mathcal B_{s_j} \cap \mathcal B_k$, and $l_2$ and $u_2$ delimit the box
$\mathcal B_k \cap \mathcal B_{s_{j+1}}$.
We compare the optimal values of the two problems
illustrated in Fig.~\ref{f-shortening}.
The first is
\[
\begin{array}{ll}
\mbox{minimize} &\|y-a\|_2 + \|b-y\|_2 \\
\mbox{subject to} &
l \leq y \leq u,
\end{array}
\]
where the only variable is $y$.
The second is
\BEQ \label{e-box-split}
\begin{array}{ll}
\mbox{minimize} &\|z_1-a\|_2 + \|z_2-z_1\|_2 + \|b-z_2\|_2 \\
\mbox{subject to} &
l_1 \leq z_1 \leq u_1, \quad l_2 \leq z_2 \leq u_2,
\end{array}
\EEQ
where the variables are $z_1$ and $z_2$.
Let $y^\star$ be the solution of the first problem, which is known to us
since we have solved~\eqref{e-shortening}.
We want to check if choosing $z_1 = z_2 = y^\star$ is
optimal for the second problem.
To do so, we look for Lagrange multipliers of problem~\eqref{e-box-split} that satisfy
complementary slackness and are dual feasible~\cite[\S5.5]{boyd2004convex}.

Complementary slackness reads
\[
\begin{array}{cc}
\lambda_1 = \frac{y^\star - a}{\|y^\star - a\|_2},
& (y^\star - l_1)^T \nu_1^+ = (y^\star - l_2)^T \nu_2^+ = 0, \\
\lambda_2 = \frac{b - y^\star}{\| b - y^\star \|_2},
& (u_1 - y^\star)^T \nu_1^- =  (u_2 - y^\star)^T \nu_2^- = 0,
\end{array}
\]
where the multipliers $\lambda_1, \; \lambda_2 \in \reals^d$ are paired with the
first and last objective terms in~\eqref{e-box-split}, $\nu_1^+, \; \nu_1^- \in \reals^d$ with the lower and upper limits in 
the first box constraint, and $\nu_2^+, \; \nu_2^-\in \reals^d$ with the second box
constraint.
The constraints of the dual of problem~\eqref{e-box-split} are
\[
\begin{array}{c}
\nu_1^+, \; \nu_1^-, \; \nu_2^+, \; \nu_2^- \geq 0, \\
\|\lambda\|_2, \; \|\lambda_1\|_2, \;  \|\lambda_2\|_2\leq 1, \\
\lambda -  \lambda_1 + \nu_1^+ + \nu_1^- = \lambda_2 - \lambda + \nu_2^+ + \nu_2^- = 0,
\end{array}
\]
where the multiplier $\lambda \in \reals^d$ is paired with the second cost
term in~\eqref{e-box-split}.

We let $L_1$ and $U_1$ be the matrices that select the entries where $l_1 < y^\star$
and $y^\star < u_1$, respectively.
We let $L_2$ and $U_2$ be defined similarly but for the limits $l_2$ and $u_2$.
After a few manipulations, the two sets of conditions above reduce to the
inequalities in~\eqref{e-box-update}.
The only variable is $\lambda$, since the values of $\lambda_1$ and $\lambda_2$ are fixed
(and known) by the complementary slackness conditions.

Finally, we observe that the norm of the Lagrange multiplier $\lambda^\star$
in~\eqref{e-lambda} can be interpreted as the elastic
force exchanged between the
points $z_1$ and $z_2$ in Fig.~\ref{f-shortening},
and is indicative of the cost decrease that we incur by
letting these points separate.
This motivates our heuristic of inserting the box for which
the vector $\lambda^\star$ has largest norm.
 
\end{document}